\documentclass[preprint,12pt]{elsarticle}

\usepackage{amssymb}
\usepackage{lipsum}

\usepackage{graphicx}%
\usepackage{float}
\usepackage{multirow}%
\usepackage{amsmath,amsfonts}%
\usepackage{amsthm}%
\usepackage{mathrsfs}%
\usepackage[title]{appendix}%
\usepackage{xcolor}%
\usepackage{textcomp}%
\usepackage{manyfoot}%
\usepackage{booktabs}%
\usepackage{algorithm}%
\usepackage{algorithmicx}%
\usepackage{algpseudocode}%
\usepackage{listings}%
\usepackage{adjustbox}
\usepackage{subcaption}%
\usepackage{graphicx}%
\usepackage{multirow}

\usepackage{caption}

\usepackage{multirow}
\newcommand\MyBox[2]{
	\fbox{\lower0.75cm
		\vbox to 1.3cm{\vfil
			\hbox to 1.3cm{\hfil\parbox{0.5cm}{#1}\hfil}
			\vfil}%
	}%
}
\usepackage{tikz}
\usetikzlibrary{positioning, fit, arrows.meta, shapes}
\usepackage{rotating}

\begin{document}

\begin{frontmatter}



\title{Some variation of COBRA in sequential learning setup}

\author[first]{Aryan Bhambu}
\ead{a.bhambu@iitg.ac.in}
\affiliation[first]{organization={Department of Mathematics, Indian Institute of Technology}, city={Guwahati}, postcode={781039}, state={Assam}, country={India}}

\author[second]{Arabin Kumar Dey\corref{cor1}} 
\ead{arabin@iitg.ac.in}
\affiliation[second]{organization={Department of Mathematics, Indian Institute of Technology}, city={Guwahati}, postcode={781039}, state={Assam}, country={India}}
 
\cortext[cor1]{Corresponding author at: Department of Mathematics, Indian Institute of Technology, India}


\pagebreak
\setcounter{page}{0}
\pagenumbering{arabic}

\begin{abstract}
This research paper introduces innovative approaches for multivariate time series forecasting based on different variations of the combined regression strategy.  We use specific data preprocessing techniques which makes a radical change in the behaviour of prediction.  We compare the performance of the model based on two types of hyper-parameter tuning Bayesian optimisation (BO) and Usual Grid search.  Our proposed methodologies outperform all state-of-the-art comparative models. We illustrate the methodologies through eight time series datasets from three categories: cryptocurrency, stock index, and short-term load forecasting.
\end{abstract}


\end{frontmatter}

\section{Introduction}

  Multivariate time series forecasting is essential as it considers the interdependencies among multiple variables, providing a more accurate understanding of complex systems \cite{de2011forecasting}. This approach proves valuable in the fields such as finance, economics, and environmental monitoring, allowing organisations to make more informed decisions by capturing underlying relationships between variables.    
  
  Researchers used not just deep learning techniques to predict future values in both univariate and multivariate time series, but also created elegant framework where they accommodated large number of machine learning models which are specifically meant for independent datasets.   Multiple researchers used different regression based models like Multiple Linear regression, Ridge, LASSO, Support Vector Regression (SVR) (\cite{tay2001application}), Decision tree based approaches to forecast time-series of various nature.   
  
  Deep Neural Networks (DNN), particularly recurrent neural networks like Long Short-Term Memory (LSTM), Gated Recurrent Unit (GRU), and Highway Long Short-Term Memory (Highway-LSTM), gained prominence due to their powerful generalisation and scalability \cite{hochreiter1997long, chung2014empirical, zilly2017recurrent}.  Literature contains even comparative studies between traditional methods like ARIMA and advanced neural network models such as LSTM and GRU for time series forecasting \cite{yamak2019comparison, fu2016using}. These studies demonstrate the superiority of certain deep learning models, such as GRU, in specific forecasting tasks like traffic time series forecasting \cite{fu2016using}. Hybrid time series forecasting models integrate diverse techniques, combining statistical methods and machine learning to enhance accuracy and adaptability, capturing both linear and non-linear patterns in the data \cite{fathi2019time, islam2021foreign}. Fathi proposed a hybrid model that merges ARIMA and LSTM for sales forecasting \cite{fathi2019time}. Islam et al. \cite{islam2021foreign} introduced a hybrid LSTM model composed of GRU and LSTM for time series forecasting, demonstrating its superiority over alternative models.   \cite{zhou2021informer} proposed a transformer-based model for long sequence time-series forecasting and fully utilise the self-attention mechanism to deal with super long sequences, outperforming the existing models on four large-scale datasets.

   Researchers over the years constantly focused on improving the accuracy by the introducing many ensemble algorithms.   K-NN ensemble \cite{martinez2019methodology} leverages the collective insights of its nearest neighbours to make predictions, effectively capturing non-linear relationships in time series data. Similarly, AdaBoost \cite{wang2021adaboost} sequentially combines weak learners to enhance overall predictive accuracy, while XGBoost [ \cite{tay2001application}, \cite{fan2018comparison}, \cite{vuong2022stock}], a powerful ensemble method, excels in capturing complex temporal dependencies through boosting.  However, these ensembles use regression based weak learners.  A framework for ensemble learning methods with sequential learning based Deep Neural Networks as weak learners are not available in the literature.
 
 Distance-based ensembles are important as they enhance model robustness by leveraging diverse algorithms, reducing overfitting, and improving generalisation across varying data patterns. By considering the dissimilarity between instances, these ensembles excel at capturing complex relationships in data, leading to more reliable and accurate predictions \cite{juditsky2000functional, yang2000combining, yang2004aggregating}. We propose a non-linear way of combining estimators, adding to a streamline of works pioneered by Mojirsheibani \cite{mojirsheibani1999combining}. Our methods Dynamic Proximity Ensemble (DPE) model extends the COBRA (Combined Regression Strategy) algorithm introduced by Biau et al. \cite{biau2016cobra}.  There is no work available based on COBRA in sequential learning framework so far.
 
  In the domain of time series forecasting, which involves preprocessing, sequence modelling, independent forecasting, and aggregation, tuning numerous hyper-parameters is essential for accurate predictions. While grid search is commonly employed due to its simplicity, it can be time-consuming and biased without correlation.   The motivation behind applying the Bayesian Optimisation Algorithm (BOA) \cite{snoek2012practical} in hyper-parameter tuning lies in controlling time complexity and preserving balance in exploitation and exploration.  The effectiveness of BOA in improving hyper-parameter tuning for various prediction models, including attention-based LSTM, PM2.5 prediction, and COVID-19 prediction are available in different research articles [\cite{alizadeh2021novel}, \cite{abbasimehr2021prediction}, and \cite{ma2020lag}].  These works validate the superiority of BOA over baseline models in achieving optimised model performance \cite{alizadeh2021novel,abbasimehr2021prediction,ma2020lag}.   The paper explores the implementation of BOA in all proposed variations and highlight the best model in this context.

 The organisation of the paper goes as follows.  Section 2 demonstrates the all proposed methodologies adapted in this sequential learning setup.  Section 3 describes implementation structure of Bayesian Optimisation for Hyper-parameter Tuning on the proposed setup.  Details of the framework for empirical studies on different datasets are available in Section 4.  We discuss the empirical results in Section 5.  Section 6 describes dynamic prediction in this context.  We conclude the paper in Section 7.

\subsection{Contributions of our proposed model}

The main contribution of the paper are:

\begin{enumerate}
    \item The paper would be first contribution of COBRA in sequential learning setup in multi-dimensional signal.
    \item We explore two different variation of COBRA, Dynamic Proximity Ensemble and Partitioned Dynamic Proximity Ensemble respectively for multivariate time series forecasting.
    \item We observe Dynamic Proximity Ensemble outperforms all other variations of COBRA and existing state or the models with respect to the two metrics considered in this paper. 
    \item We apply the computationally efficient Bayesian optimisation algorithm to automatically search the optimal ensemble configurations.
    \item We experiment with the eight different time series datasets having three groups of datasets:  Cryptocurrency, Stock Index, and Australian Energy Market Operator. The computational and statistical experiments evidence superiority of our proposed model over others.
\end{enumerate}

\section{Proposed Methodologies}
\subsection{Dynamic Proximity based Ensemble (DPE)}

Let us assume the dataset $D_{ori}=\{ (Y_{11}, \cdots Y_{N1}), \ldots , (Y_{1T}, \cdots, Y_{NT}) \}$ and $ D_{ori} = (\underline{Y}_{1}, \cdots, \underline{Y}_{T}) \in \mathbb{R}^N$.  The initial objective is to make one step ahead prediction in a time-series data or to find out the expectation of $\underline{Y}_{T}$ given $\mathcal{F}_{T - 1} = \{ \underline{Y}_{1}, \underline{Y}_{2}, \cdots, \underline{Y}_{T -1} \}$.  
  
 The current form of COBRA available in the literature so far is valid for independent datasets which is meant for usual regression purpose or with an extension in functional regression (Goswami et al. ).

  We propose the following initial steps to transfer the time series problems into a similar framework/design that can mimic the existing COBRA.   We use sliding window to create the frames.   It will help to transfer the whole data into series of exchangeable datasets with a specific window length of covariates and a target variable to perform COBRA on these exchangeable sets.

In multidimensional data, we are creating a tensor ($D_1$) where each matrix will consist of exchangeable set of elements.    Thus, we can use COBRA in multi-dimensional time series framework.

 Let us denote the i-th frame/matrix with $l$ time step of the tensor is 
 
 $F^{l}_{i} = 
\begin{bmatrix}
    Y_{1, i}       & Y_{1, i + 1} & Y_{1, i + 2} & \dots & Y_{1, i + l} \\
    Y_{2, i}       & Y_{2, i + 1} & Y_{2, i + 2} & \dots & Y_{2, i + l} \\
    \hdotsfor{5} \\
    Y_{N, i}       & Y_{2, i + 1} & Y_{2, i + 2} & \dots & Y_{N, i + l} \\
\end{bmatrix}
$
\\
Thus,  $D_1 = \{ F^{l}_{1}, \cdots, F^{l}_{T - l + 1} \}$.  

We further create a new set of exchangeable sets taking both frames of covariates and the target observations.  We call the set as $D^{cob}$.
 Therefore, $D^{cob} = \{ (F^{l}_{1}, \underline{Y}_{1}), \cdots, (F^{l}_{T - l + 1} , \underline{Y}_{T - l + 1})  \}$.

According to our proposition, we divide $D^{cob}$ into train, validation and test set.  For better understanding of the methodology, let us consider training set $D_n =\{ (F^{l}_{1}, \underline{Y}_{1}), \cdots, (F^{l}_{n}, \underline{Y}_{n}) \}$ and validation set \\
$D_v = \{ (F^{l}_{n + 1}, \underline{Y}_{n + 1}) \cdots, (F^{l}_{n + v}, \underline{Y}_{n + v}) \}$.

In the DPE framework, $M$ competing estimators, referred to as machines, are employed. Each machine $m$ is trained on $D_n$ and denoted as $r_{tr}^{1}, r_{tr}^{2}, \ldots , r_{tr}^{M}$, with $r_{tr}^{i}$ representing a machine trained exclusively on $D_n$ (train) capable of estimating $E( \underline{Y}_{t} | F^{l}_{t - l - 1} = x ) $ for any $x \in \mathbb{R}^{t -1}$. The DPE estimate of $E( \underline{Y}_{t} | F^{l}_{t - l - 1} = x ) $ is then formulated as :

\begin{equation}
    \hat{E}_{DPE}( \underline{Y}_{t} | F^{l}_{t - l - 1} = x ) = \sum_{i=1}^{n} W_{n, i}(x) \underline{Y}_{i}     \label{dpeeqn1}
\end{equation}

The weight $W_{n,i}(x)$ for the $i$-th sample in $D_n$ is defined by:
\begin{equation}
    W_{n,i}(x) = \frac{I\left(\bigcap_{m=1}^{M} || r_{tr}^{m}(F^{l}_i) - r_{tr}^{m}(x) || \leq \epsilon \right)}{\sum_{j=1}^{n} I\left(\bigcap_{m=1}^{M} || r_{tr}^{m}(F^{l}_j) - r_{tr}^{m}(x) || \leq \epsilon \right)}     \label{dpeeqn2}
\end{equation}

Here, $\epsilon$ is distance threshold, a user-specified parameter. The weight reflects the consensus across machines regarding the proximity of the $i$th frame to the query frame $x$. Specifically, if the $i$th frame is close to $x$ in all machines, the weight is 1; otherwise, if it is distant in any machine, the weight is 0. A new parameter, $\alpha$, determines the minimum number of concurring machines for the weight to be 1, thus introducing an element of consensus. The weight can be expressed as:

\begin{equation}
    W_{n,i}(x) = \frac{I\left(\sum_{m=1}^{M} |I \left(| r_{tr}^{m}(F^{l}_i) - r_{tr}^{m}(x) || \leq \epsilon \right) \geq M\alpha \right)}{\sum_{j=1}^{n} I\left(\sum_{m=1}^{M} I\left(|| r_{tr}^{m}(F^{l}_j) - r_{tr}^{m}(x) || \leq \epsilon \right) \geq M\alpha \right)}     \label{dpeeqn3}
\end{equation}

\begin{figure*}[!ht]
	\centering
	\includegraphics[scale=0.4]{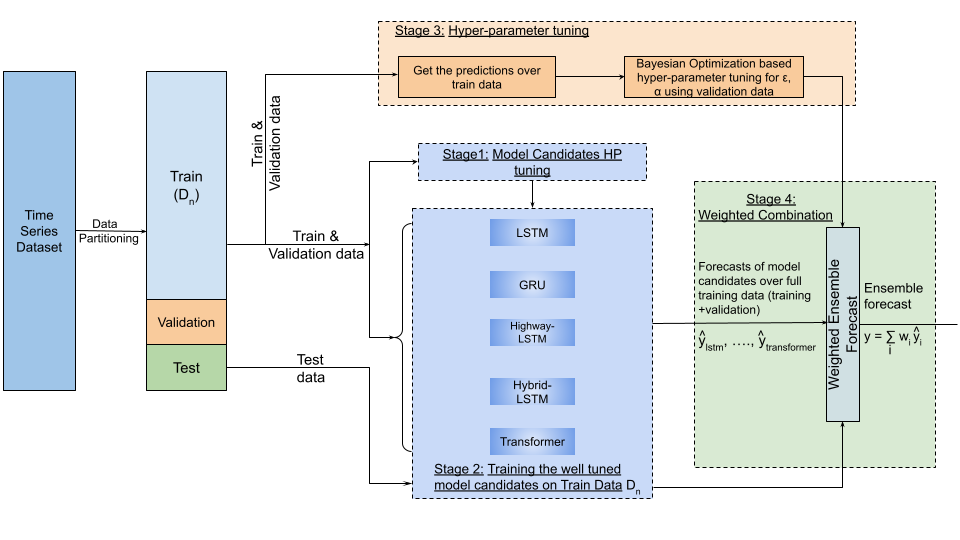}
	\label{fig:srpadpe}
	\caption{The schematic representation of the Dynamic Proximity Ensemble (DPE) approach. It condenses the model into four key phases: fine-tuning of model candidates' hyper-parameters, configuring combinations via the Bayesian optimisation algorithm (BOA), retraining the model candidates, generating the ensemble output with the optimal combination configuration.}
\end{figure*}

\subsection{Partition-Dynamic Proximity based Ensemble (PaDPE)}

PaDPE also initially starts with three division of the data : training, validation and test.  However, we divide further the training dataset $D_n$ into two distinct parts in PaDPE : $D_{n_1}= (F^{l}_{1}, \underline{Y}_{1}), \ldots (F^{l}_{n_1}, \underline{Y}_{n_1})$ and $D_{n_2}= (F^{l}_{n_1 + 1}, \underline{Y}_{n_1 + 1}), \ldots , (F^{l}_{n}, \underline{Y}_{n})$, where $n_2 = n - n_1 \geq 1$. We denote validation set as $D_v = \{ (F^{l}_{n + 1}, \underline{Y}_{n + 1}) \cdots, (F^{l}_{n + v}, \underline{Y}_{n + v}) \}$.

In the PaDPE framework, $M$ competing estimators, referred to as machines, are employed. Each machine $m$ is trained on $D_{n_1}$ and denoted as $r_{n_1}^{1}, r_{n_1}^{2}, \ldots , r_{n_1}^{M}$, with $r_{n_1}^{i}$ representing a machine trained exclusively on $D_{n_1}$ capable of estimating  $\hat{E}_{PaDPE}( \underline{Y}_{t} | F^{l}_{t - l - 1} = x ) $ for any $x \in \mathbb{R}^{l \times N}$. The PaDPE estimate of $ \hat{E}_{PaDPE}( \underline{Y}_{t} | F^{l}_{t - l - 1} = x )$ is then formulated as:
\begin{equation}
    \hat{E}_{PaDPE}( \underline{Y}_{t} | F^{l}_{t - l - 1} = x ) = \sum_{i=1}^{n} W_{n, i}(x) \underline{Y}_{i}     \label{padpeeqn1}
\end{equation}

The weight $W_{n,i}(x)$ for the $i$th sample in $D_{n}$ is defined by:
\begin{equation}
    W_{n,i}(x) = \frac{I\left(\bigcap_{m=1}^{M} || r_{n_1}^{m}(F^{l}_i) - r_{n_1}^{m}(x) || \leq \epsilon \right)}{\sum_{j=1}^{n} I\left(\bigcap_{m=1}^{M} || r_{n_1}^{m}(F^{l}_j) - r_{n_1}^{m}(x) || \leq \epsilon \right)}     \label{padpeeqn2}
\end{equation}

Here, $\epsilon$ is distance threshold, a user-specified parameter. The weight reflects the consensus across machines regarding the proximity of the $i$th sample to the query point $x$. Specifically, if the $i$th sample is close to $x$ in all machines, the weight is 1; otherwise, if it is distant in any machine, the weight is 0. A new parameter, $\alpha$, determines the minimum number of concurring machines for the weight to be 1, thus introducing an element of consensus. The weight can be expressed as:
\begin{equation}
    W_{n,i}(x) = \frac{I\left(\sum_{m=1}^{M} I\left(|| r_{n_1}^{m}(F^{l}_i) - r_{n_1}^{m}(x) || \leq \epsilon \right) \geq M\alpha \right)}{\sum_{j=1}^{n} I\left(\sum_{m=1}^{M} I\left(|| r_{n_1}^{m}(F^{l}_j) - r_{n_1}^{m}(x) || \leq \epsilon \right) \geq M\alpha \right)}     \label{padpeeqn3}
\end{equation} 

\begin{figure*}[!ht]
	\centering
	\includegraphics[scale=0.4]{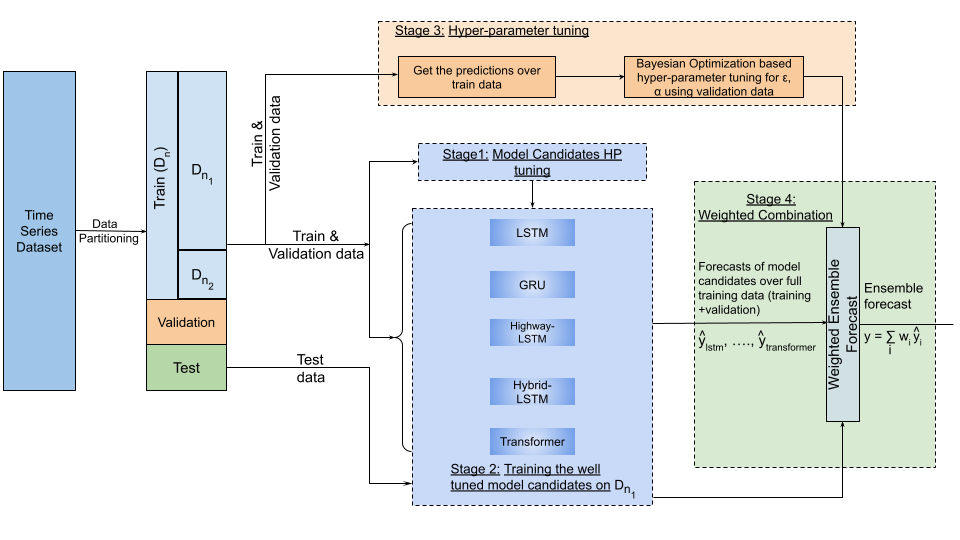}
	\label{fig:srdpe}
	\caption{The schematic representation of the Partition-Dynamic Proximity Ensemble (PaDPE) approach. It condenses the model into four key phases: fine-tuning of model candidates' hyper-parameters, configuring combinations via the Bayesian optimisation algorithm (BOA), retraining the model candidates, generating the ensemble output with the optimal combination configuration.}
\end{figure*}

\subsection{Training}

Moreover, detailed explanations regarding the training methods for the proposed algorithms are available in algorithms \ref{traindpealgo} and \ref{trainpadpealgo}. This subsection provides a comprehensive understanding of the steps and processes involved in training the models providing insights into the training procedures for both the DPE and PaDPE algorithms.

To delve into specifics, algorithms \ref{traindpealgo} and \ref{trainpadpealgo} offers a detailed walkthrough of the training process for the DPE and PaDPE algorithm, explaining the steps taken to fine-tune the model parameters and improve its predictive capabilities. This includes initiating the base models, calculating distances between predictions, and determining weights, all contributing to the generation of accurate predictions.

Furthermore, the subsequent section provides a comprehensive overview of Bayesian Optimisation. This serves as a precursor to understanding how both the DPE and PaDPE models undergo optimisation using Bayesian techniques. 

\begin{algorithm}
    \caption{Training Algorithm for DPE network}
    \label{traindpealgo}
    \begin{algorithmic}[1]
        \Statex \textbf{Input:} Training dataset $D_n$, distance threshold $\epsilon$, $M$ number of base models, and fraction of base models $\alpha$.
        \Statex \textbf{Output:} Prediction $\hat{y}$ for any test observation $x$.
        \State Train $M$ base models using the training dataset $D_n$.
        
        \For{$i \gets 1$ to $n$}
            \State Obtain predictions $r_{tr}^{m}(F^{l}_i)$ for each model $m$.
            \State Calculate the distance between $r_{tr}^{m}(F^{l}_i)$ and $r_{tr}^{m}(x)$.
        \EndFor
        \State Calculate the indicator function and weights for each training sample with respect to the test observation using $\epsilon$, and $\alpha$ from equation \ref{dpeeqn3}.
        \State After calculation of weights, calculate their product with the training labels $Y_i$, resulting in the prediction $\hat{y}$ according to Equation \ref{dpeeqn1}.
    \end{algorithmic}
\end{algorithm}

\begin{algorithm}
    \caption{Training Algorithm for PaDPE network}
    \label{trainpadpealgo}
    \begin{algorithmic}[1]
        \Statex \textbf{Input:} Training dataset $D_n$, fraction $\frac{n_1}{n}$, distance threshold $\epsilon$, $M$ number of base models, and fraction of base models $\alpha$.
        \Statex \textbf{Output:} Prediction $\hat{y}$ for any test observation $x$.
        \State Partition the training dataset $D_n$ into $D_{n_1}$, and $D_{n_2}$ using the fraction $\frac{n_1}{n}$.
        \State Train $M$ base models on the dataset $D_{n_1}$.
        
        \For{$i \gets 1$ to $n$}
            \State Obtain predictions $r_{n_1}^{m}(F^{l}_i)$ for each model $m$.
            \State Calculate the distance between $r_{n_1}^{m}(F^{l}_i)$ and $r_{n_1}^{m}(x)$.
        \EndFor
        \State Calculate the indicator function and weights for each training sample with respect to the test observation using $\epsilon$, and $\alpha$ from equation \ref{padpeeqn3}.
        \State After calculation of weights, calculate their product with the training labels $Y_i$, resulting in the prediction $\hat{y}$ according to Equation \ref{padpeeqn1}.
    \end{algorithmic}
\end{algorithm}

\subsection{Testing}

In this study, we delve into testing procedures for the DPE and PaDPE methodologies. These approaches aim to estimate the conditional expectations  $ \hat{E}_{DPE}( \underline{Y}_{t} | F^{l}_{t - l - 1} = x ) $ and $\hat{E}_{PaDPE}( \underline{Y}_{t} | F^{l}_{t - l - 1} = x )$ respectively.  Note that $\underline{Y}_{t}$ is a point from test dataset,  $F^{l}_{t - l - 1}$ is the respective query frame.   In a different note, our aim is to make one step ahead forecasting of a multidimensional time series given a new frame.  

\subsubsection{DPE Methodology on Test dataset}

For the DPE method, proximity frames are from the combined dataset comprising of predictions from both $D_n$ (training data) and $D_v$ (validation data). The DPE estimate of $\hat{E}_{DPE}( \underline{Y}_{t} | F^{l}_{t - l - 1} = x )$  is formulated as:

\begin{equation}
  \hat{E}_{DPE}( \underline{Y}_{t} | F^{l}_{t - l - 1} = x )  = \sum_{i=1}^{n} W_{n, i}(x) \underline{Y}_{i}   \label{dpeeqn4}
\end{equation}

Here, the weights $(W_{nv,i}(x))$ are computed using the following equation:

\begin{equation}
	W_{nv,i}(x) = \frac{I\left(\sum_{m=1}^{M} I\left(|| r_{tr}^{m}(F^{l}_i) - r_{tr}^{m}(x) || \leq \epsilon \right) \geq M\alpha \right)}{\sum_{j=1}^{n+v} I\left(\sum_{m=1}^{M} I\left(|| r_{tr}^{m}(F^{l}_j) - r_{tr}^{m}(x) || \leq \epsilon \right) \geq M\alpha \right)} \label{dpeeqn5}
\end{equation}

In this formulation, $r_{tr}^{m}(F_i)$ represents the m-th machine for the frame $F^{l}_i$ where $F^{l}_i$  is some arbitrary frame from combined set  $D_n$ and $D_v$ , and $\epsilon$ and $\alpha$ are threshold parameters.

\subsubsection{PaDPE Methodology}

 In PaDPE Methodology, the testing space is similar to the tuning space.   Proximity set consists of predictions from a smaller training set as well as predictions on new points in both during tuning parameters and prediction on new query points from test dataset.

The PaDPE methodology involves a more extensive proximity point set consists of predictions on a combined space of $D_{n_1}$ and $D_{n_2}$ and validation data $D_v$. The PaDPE estimate of $\hat{E}_{PaDPE}( \underline{Y}_{t} | F^{l}_{t - l - 1} = x )$ follows a similar formulation as the DPE method:

\begin{equation}
	 \hat{E}_{PaDPE}( \underline{Y}_{t} | F^{l}_{t - l - 1} = x ) = \sum_{i=1}^{n} W_{n, i}(x) \underline{Y}_{i}  \label{padpeeqn4}
\end{equation}

where, $F^{l}_{t - l - 1} = x$ is the query frame from test dataset.

The corresponding weight calculation is given by:

\begin{equation}
	W_{nv,i}(x) = \frac{I\left(\sum_{m=1}^{M} I\left(|| r_{n_1}^{m}(F^{l}_i) - r_{n_1}^{m}(x) || \leq \epsilon \right) \geq M\alpha \right)}{\sum_{j=1}^{n+v} I\left(\sum_{m=1}^{M} I\left(|| r_{n_1}^{m}(F^{l}_j) - r_{n_1}^{m}(x) || \leq \epsilon \right) \geq M\alpha \right)} \label{padpeeqn5}
\end{equation}

In this context, $r_{n_1}^{m}(F^{l}_i)$ denotes the m-th machine for the frame $F^{l}_{i}$ from a combined space of $D_{n_1}$ and $D_{n_2}$ and $D_v$.  Note that, $\epsilon$ and $\alpha$ remain the threshold parameters.

\section{Bayesian Optimisation for Hyperparameter Tuning}

Multiple hyper parameter together creates large number of possible models.  In automated learning one popular technique to select the best model is Bayesian Optimisation which selects the hyper-parameter based on on a surrogate function to model the conditional probability of performance on a validation set. Unlike traditional grid search methods, Bayesian Optimisation Algorithm (BOA) preserves and leverages all prior computations, avoiding redundant evaluations of suboptimal hyper-parameter configurations. The algorithm employs an acquisition function to systematically identify the most promising hyper-parameter configuration for evaluation in subsequent iterations.

The Bayesian Optimisation Algorithm (BOA) comprises five key components: the surrogate function, the hyper-parameter search space, the acquisition function, the objective function, and the historical record of evaluations. In this context, the objective function is defined as the predictive performance on the validation set. The surrogate function is established using the Tree-based Parzen Window Estimation (TPE) algorithm, and the selected acquisition function is the expected improvement. The formulation of the acquisition function, denoted as \( S_{f^{\star}} (v) \), is expressed as follows:

\begin{equation}
    S_{f^{\star}} (v) = \int_{-\infty}^{f^{\star}} (f^{\star} - f) P(f | v) \, df. \label{BOE}
\end{equation}

Here, \( f^{\star} \) represents the objective function, and \( f \) represents the predefined threshold for the objective function concerning the hyper-parameters \( v \). The Tree-based Parzen Estimation (TPE) algorithm encompasses several steps, as elucidated in Algorithm \ref{BOEalgo}.

\begin{algorithm}
    \caption{TPE for Hyperparameter Tuning}
    \label{BOEalgo}
    \begin{algorithmic}[1]
        \Statex \textbf{Input:} Objective function \( f \), TPE method \( \mathcal{M} \), hyper-parameter space \( \mathbb{H}_{v} \), acquisition function validation \( \mathcal{S} \), initialised memory \( \mathcal{D} \).
        
        \Statex \textbf{Output:} Best hyperparameters \( v^{\star} \) in memory \( \mathcal{D} \).
        
        \For{\( i \gets 1 \) to \( N \)}{
            \State \( P(f | v) \) $\gets$ Fit memory \( \mathcal{D} \) using \( \mathcal{M} \)
            \State \( v_{i+1} \) $\gets$ Maximize acquisition function \( \mathcal{S} \) in Equation 
            \Statex \hspace{0.5cm}(\ref{BOE}) to determine the next hyper-parameter choice
            \State \( f(v_{i+1}) \) $\gets$ Evaluate the objective function
            \State \( \mathcal{D} \) $\gets$ \( \mathcal{D} \) $\cup$ \( (v_{i+1}, f(v_{i+1})) \)
        \EndFor}
    \end{algorithmic}
\end{algorithm}

This algorithm iteratively refines hyper-parameter choices, leveraging the acquisition function to guide the search toward promising regions within the hyper-parameter space. The historical record \( \mathcal{D} \) accumulates optimal configurations, ultimately leading to improved model performance.

 For our proposed methodology, The objective function aims to minimise the mean squared error. The search space defines hyper-parameter values. The Tree of Parzen Estimators (TPE) algorithm, used implicitly in hyperopt, guides the exploration-exploitation trade-off through its acquisition function, suggesting where to sample for optimal hyper-parameters.

\begin{figure*}[!ht]
	\centering
	\includegraphics[scale=0.27]{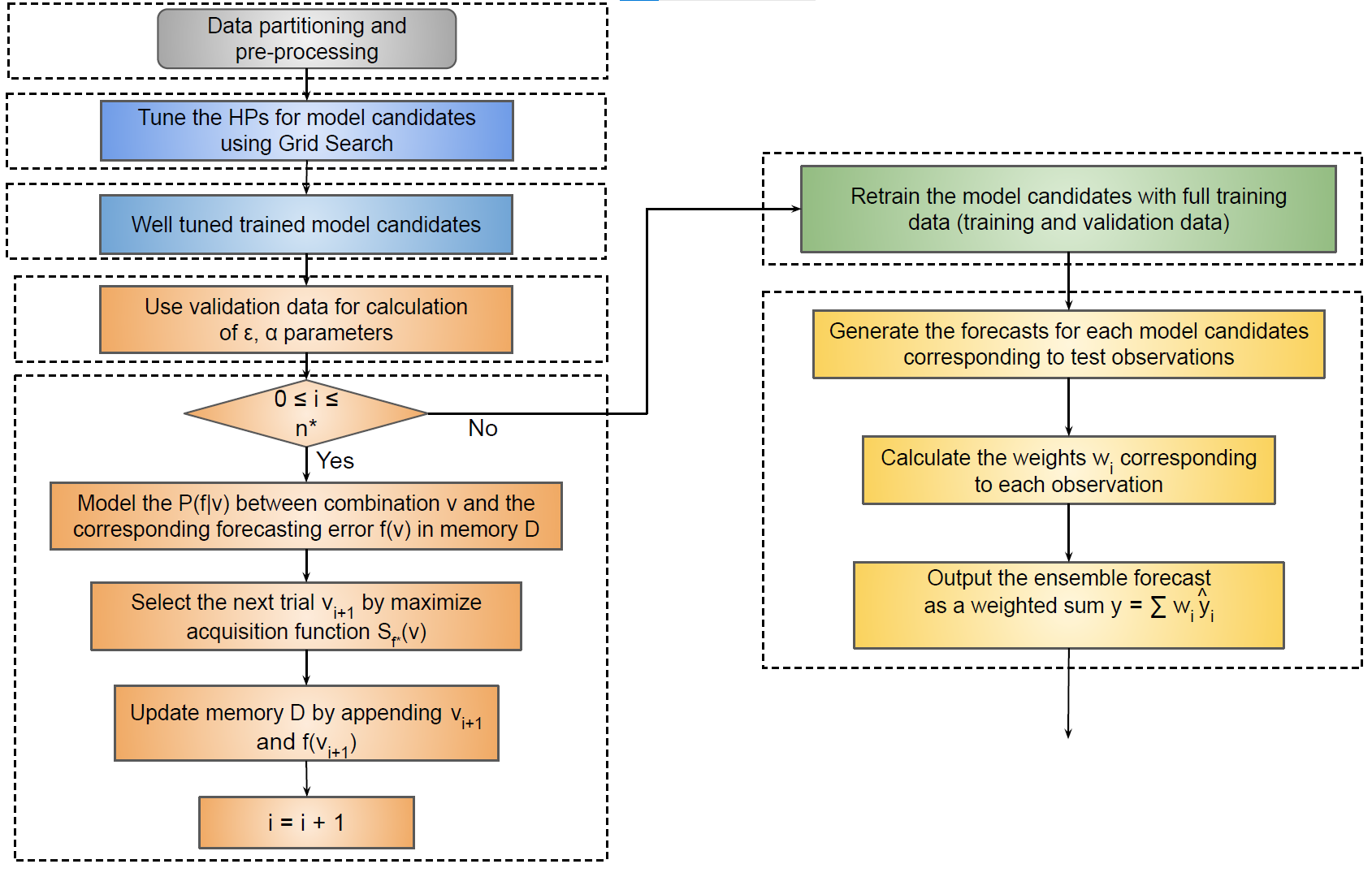}
	\label{fig:fcdpe}
	\caption{The schematic flowchart of the proposed dynamic proximity ensemble forecasting framework}
\end{figure*}

\section{Empirical Study}

This section presents the empirical study conducted on eight time series datasets, comprising cryptocurrency, stock index, and load forecasting time series datasets. The cryptocurrency and stock index finance datasets have been gathered from Yahoo Finance, while the load time series forecasting datasets have been obtained from the Australian Energy Market Operator (AEMO). Initially, we present a brief introduction to the datasets and pre-processing steps.   We also describe the assessment metrics, benchmark models, and hyper-parameter optimisation in this context.   

\subsection{Data and its nature}

  The first three, covering the period from January $1, 2015$, to June $30, 2023$, are financial datasets from Yahoo Finance. The Bitcoin dataset represents the cryptocurrency's daily market performance. The DJI dataset tracks the Dow-Jones Industrial Average, reflecting the performance of $30$ major U.S. companies. The S\&P$500$ dataset covers $500$ large-cap American companies, providing a comprehensive view of the U.S. stock market. All datasets are publicly accessible on Yahoo Finance's official website.  Summary of the dataset is available in Table-\ref{fin-dataset-summary}.

\begin{figure*}
	\centering
	\begin{subfigure}[b]{0.44\textwidth}
		\includegraphics[width=\textwidth]{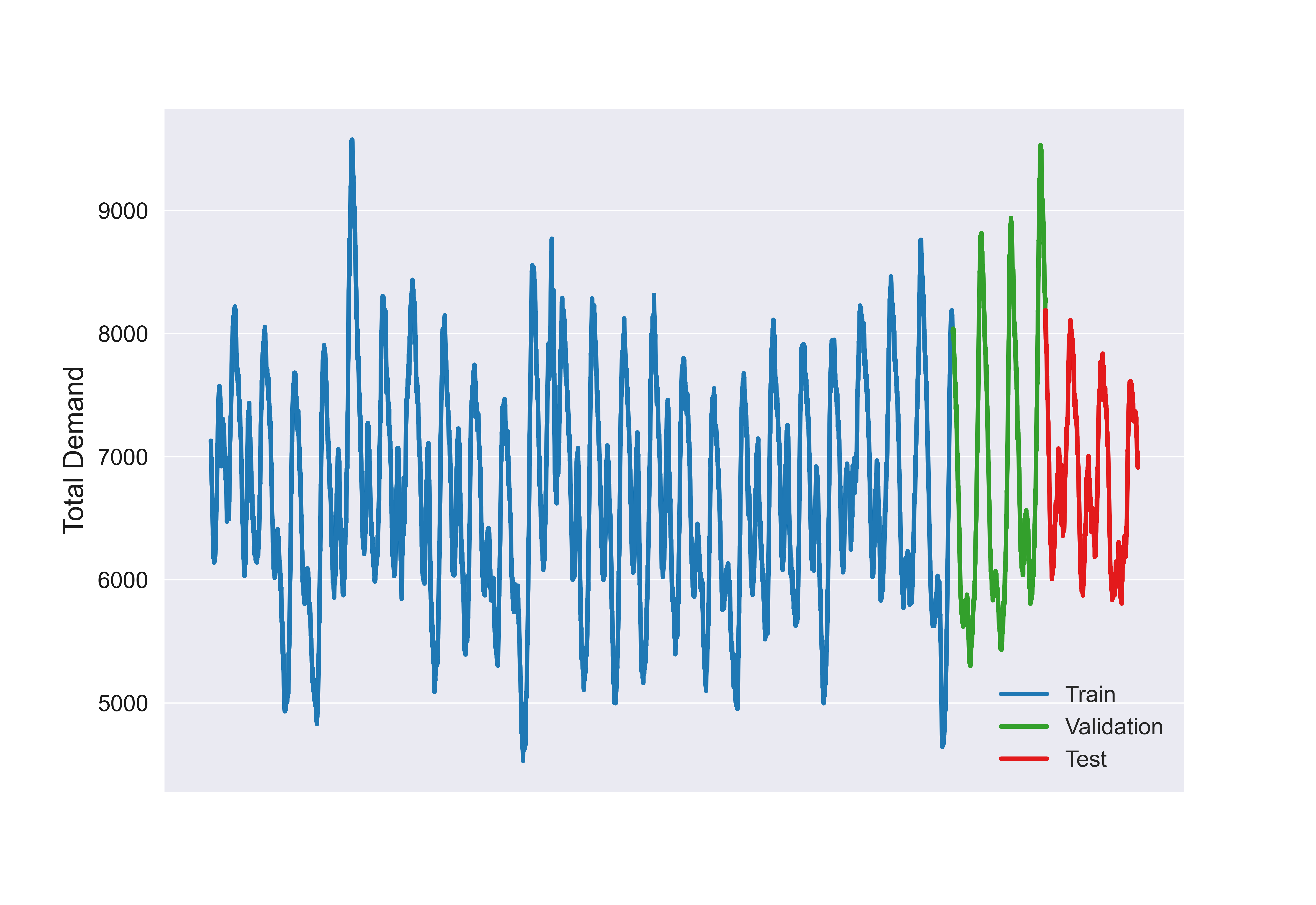}
		\caption{Total Demand - Train, Validation, and Test partitions of the raw data short-term load forecast time series data for NSW.}
	\end{subfigure}
	\hspace{0.1\textwidth}
	\begin{subfigure}[b]{0.44\textwidth}
		\includegraphics[width=\textwidth]{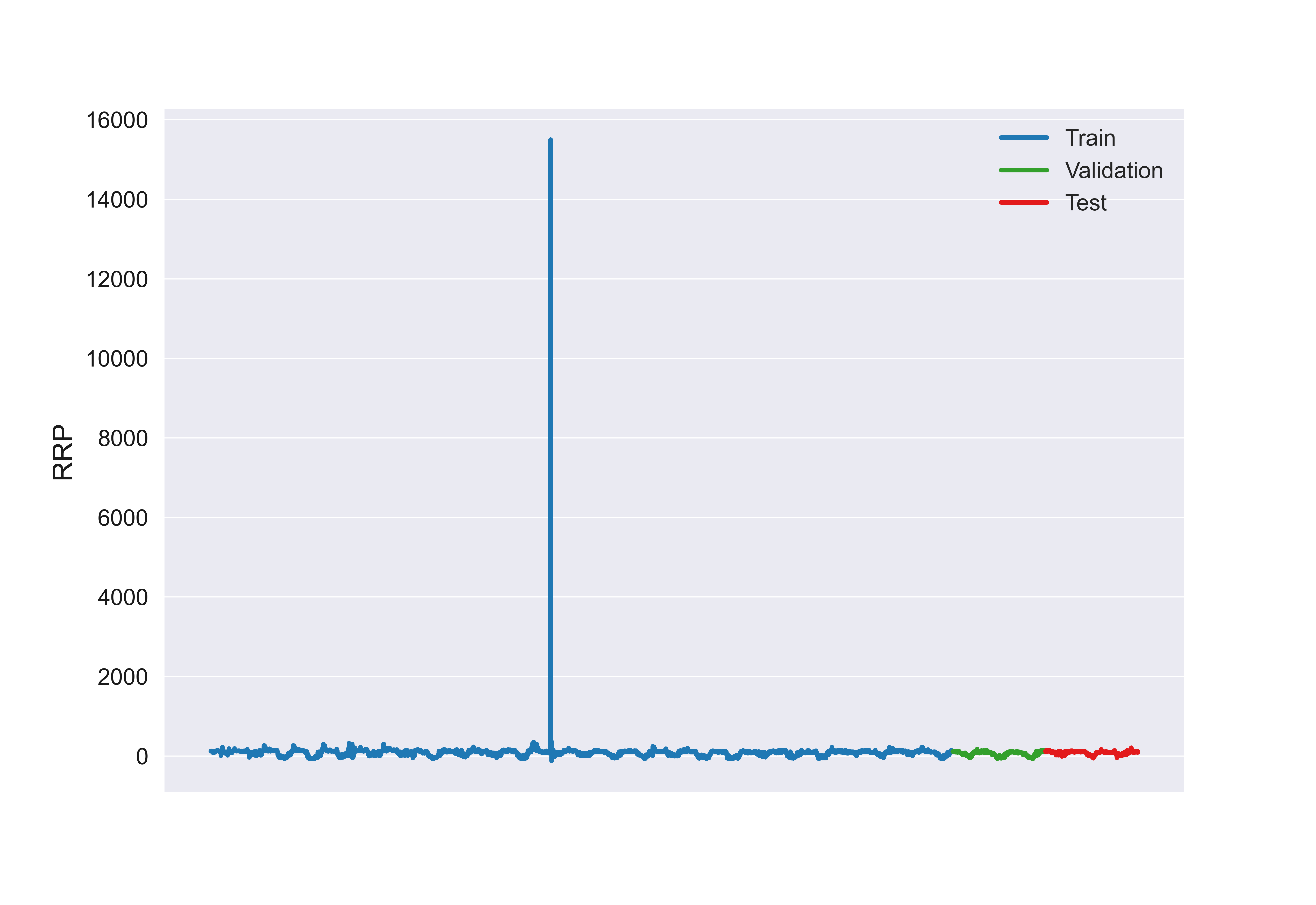}
		\caption{RRP - Train, Validation, and Test partitions of the raw data short-term load forecast time series data for NSW.}
		\label{fig:NSWRRP}
	\end{subfigure}
	\caption{Short-term load forecast time series data for New South Wales (NSW) showing the train, validation, and test partitions of the raw data for total demand and RRP respectively.}
	\label{fig:NSWdata}
\end{figure*}

\begin{figure*}
	\centering
	\begin{subfigure}[b]{0.44\textwidth}
		\includegraphics[width=\textwidth]{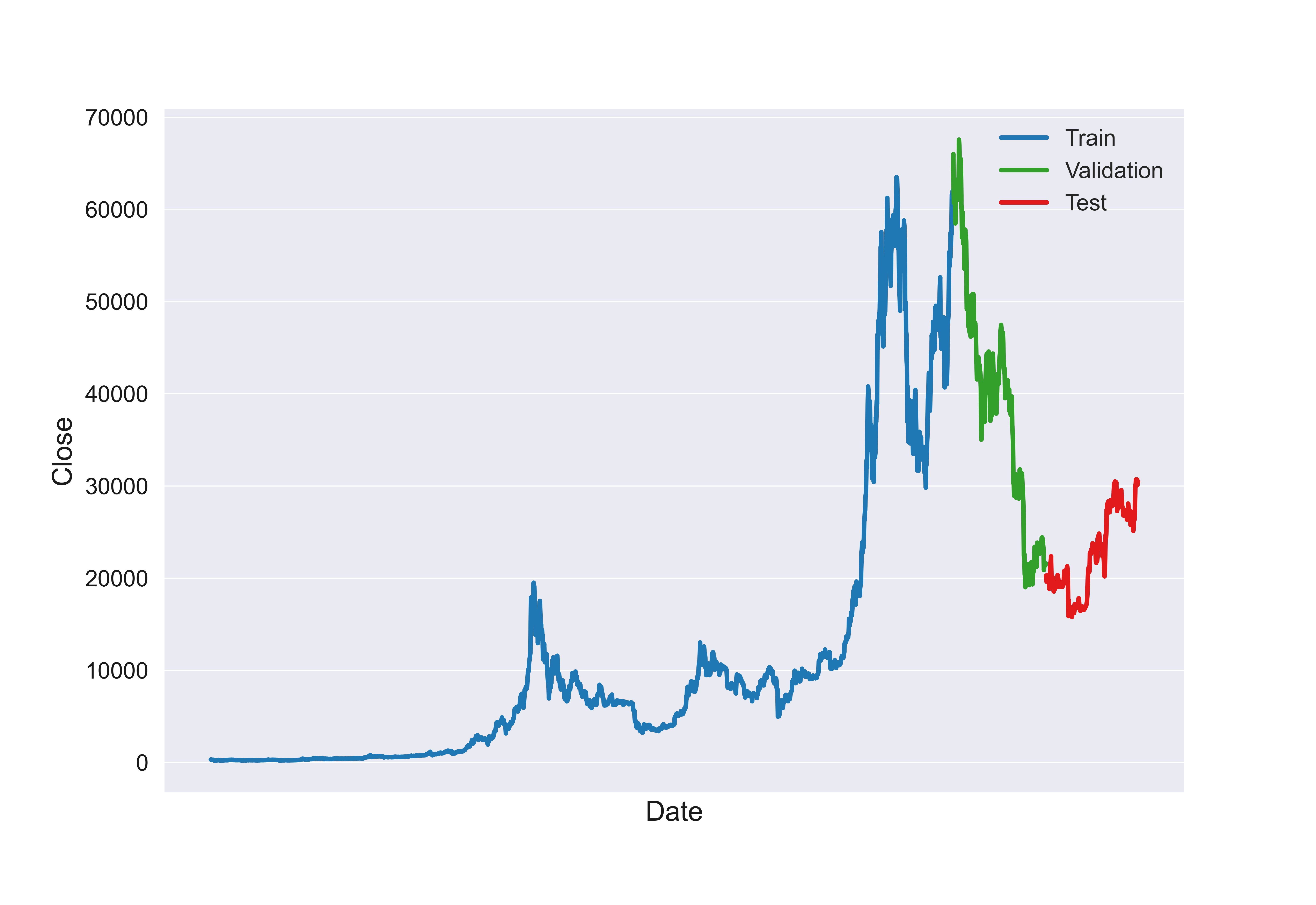}
		\caption{Closing Price - Train, Validation, and Test partitions of the raw data time series for Bitcoin cryptocurrency.}
	\end{subfigure}
	\hspace{0.1\textwidth}
	\begin{subfigure}[b]{0.44\textwidth}
		\includegraphics[width=\textwidth]{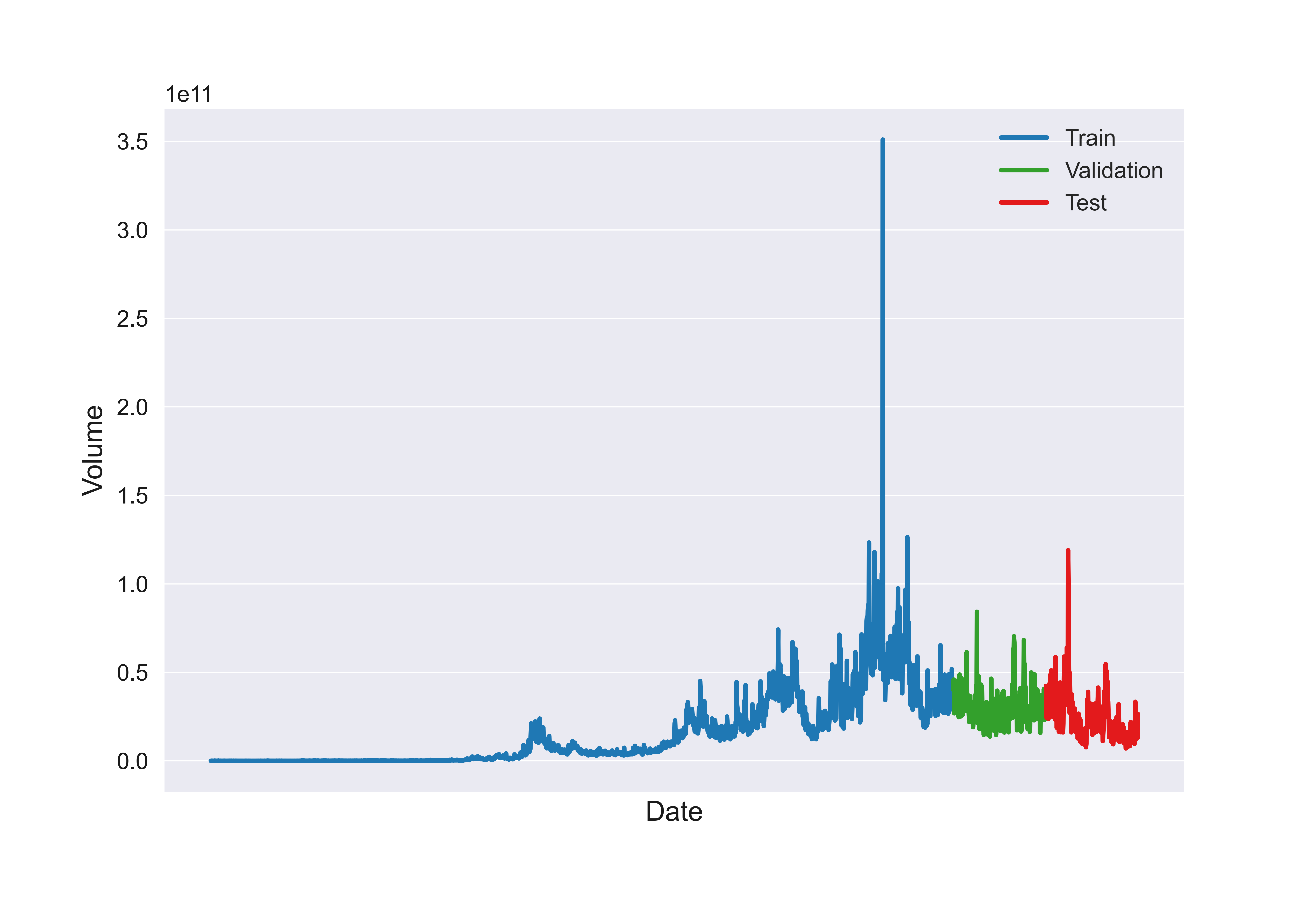}
		\caption{Volume - Train, Validation, and Test partitions of the raw data time series for Bitcoin cryptocurrency.}
		\label{fig:BitV}
	\end{subfigure}
	\caption{Time series data for Bitcoin crypto-currency, illustrating the train, validation, and test partitions of the raw data for closing price and volume.}
	\label{fig:bitcoindata}
\end{figure*}

  The subsequent five datasets encompass load data collected from the states of South Australia (SA), New South Wales (NSW), Victoria (VIC), and Tasmania (TAS) for the month of December $2022$.  The records of data are available for every half an hour, resulting in $48$ data points per day. These datasets are also publicly available and can be accessed on the official website of the Australian Energy Market Operator.  Table-\ref{load-dataset-summary} provides a brief summary of load datasets.
 We also show the data visualisation for the Bitcoin data: Closing and Volume Price and for NSW data: Total Demand and RRP in the figures: \ref{fig:NSWdata} and \ref{fig:bitcoindata} respectively.
  
\begin{table*}[!htbp]
    \centering
    \caption{Dataset Summary for the Closing price and Volume for the finance datasets}
    \label{fin-dataset-summary}
    \begin{adjustbox}{width=\textwidth}
        \begin{tabular}{|c|c|c|c|c|c|c|c|c|c|}
        
            \hline
            Dataset Name & Max & Min & Median & Mean & Std & Skewness & Kurtosis \\
            \hline
            
            Bitcoin & 67566.82 & 178.10 & 8041.77 & 14024.54 & 16089.06 & 1.36 & 0.88 \\
            DJI & 36799.64 & 15660.17 & 25669.32 & 25783.14 & 6059.98 & 0.06& -1.23 \\
            S\&P $500$ & 4796.56 & 1829.07 & 2842.73 & 3046.08 & 830.93 & 0.42 & -1.13 \\
            \hline
            
            Bitcoin & 350.9e+9 & e+9 & 0.07e+9 & 17.1e+9 & 19.6e+9 & 2.73 & 28.28 \\
            DJI & 0.9e+9 & 0.04e+9 & 0.2e+9 & 0.2e+9 & 0.1e+9 & 0.57 & 1.34 \\
            S\&P $500$ & 9.9e+9 & 1.2e+9 & 3.8e+9 & 4e+9 & 0.9e+9 & 1.69 & 4.87 \\
            \hline
    
        \end{tabular}
    \end{adjustbox}
\end{table*}

\begin{table*}[!htbp]
    \centering
    \caption{Dataset Summary for the Closing price and Total demand of the finance and load time series datasets respectively}
    \label{load-dataset-summary}
    \begin{adjustbox}{width=\textwidth}
        \begin{tabular}{|c|c|c|c|c|c|c|c|c|c|}
        
            \hline
            Month & Max & Min & Median & Mean & Std & Skewness & Kurtosis \\
            \hline
            
            SA & 2834.54 & 119.6 & 1240.97 & 1135.00 & 419.97 & -0.16 & 0.47\\
            NSW & 9575.59 & 4530.54 & 6636.68 & 6721.05 & 902.91 & 0.25 & -0.42 \\
            VIC & 8148.52 & 1976.53 & 4194.06 & 4239.37 & 788.26 & 0.54 & 2.18 \\
            TAS & 1398.93 & 770.52 & 1079.82 & 1082.54 & 92.08 & 0.18 & -0.26 \\
            QLD & 8873.19 & 3859.83 & 5497.31 & 5736.49 & 924.75 & 0.73 & 0.23 \\
            \hline
            
            SA & 12523.61 & -1000.0 & 23.08 & 35.43 & 180.14 & 42.41 & 2701.58\\
            NSW & 15500.0 & -120.0 & 93.35 & 82.64 & 179.12 & 72.52 & 6168.07 \\
            VIC & 316.21 & -937.19 & 19.15 & 31.28 & 71.74 & -0.49 & 7.96 \\
            TAS & 15493.4 & -51.65 & 83.85 & 76.35 & 169.75 & 84.00 & 7617.79 \\
            QLD & 1250.0 & -122.19 & 92.77 & 85.46 & 80.11 & 1.83 & 15.90 \\
            \hline
    
        \end{tabular}
    \end{adjustbox}
\end{table*}  

We adapt a set of specific data pre-processing techniques to ensure that machine learning models yield precise outputs. In our methodology, first we employ the max-min normalisation technique to refine the raw data. We assume that the highest and lowest values within the training set represented by \(x_{\text{max}}\) and \(x_{\text{min}}\), respectively. The process transforms the data values in [0,1] using the following equation:
\[
x_{\text{normalized}} = \frac{x - x_{\text{min}}}{x_{\text{max}} - x_{\text{min}}}
\]

Here, \(x_{\text{normalized}}\) represents the normalised time series, and \(x\) denotes the original time series.

  To implement cross-validation, we first divide the dataset into three parts: the training set, validation set, and test set.  We choose the validation and test sets each with 10\% of the entire dataset.  Furthermore, the dataset underwent cumulative sum (cumsum) pre-processing before being subjected to scaling.


\subsection{Assessment Metrics}
\subsubsection{Root Mean Square Error (RMSE)}

 RMSE measures the average magnitude of the errors between predicted ($\hat{y}_i$) and actual ($y_i$) values. Squaring the differences emphasizes larger errors, and taking the square root provides an interpretable metric in the same unit as the target variable. RMSE is sensitive to outliers due to the squaring effect.

\begin{equation}
	\text{RMSE} = \sqrt{\frac{\sum_{i=1}^{n}(y_i - \hat{y}_i)^2}{n}}
\end{equation}

\subsubsection{Mean Absolute Percentage Error (MAPE)}

 MAPE represents the average percentage difference between predicted ($\hat{y}_i$) and actual ($y_i$) values. It is particularly useful when dealing with variables of different scales, as it normalizes the error by the magnitude of the actual values. However, MAPE can be problematic when the actual values are close to zero.

\begin{equation}
	\text{MAPE} = \frac{1}{n} \sum_{i=1}^{n}\left(\frac{|y_i - \hat{y}_i|}{|y_i|}\right) \times 100
\end{equation}

\subsection{Base Models}

 In our study, we have employed various deep learning models as the foundation or base models, also referred to as learners. These models are selected for their proficiency in capturing intricate patterns and dependencies within data. The following are the base models utilised in our research:

\begin{enumerate}
	\item \textbf{Long Short-Term Memory (LSTM):} LSTM is a type of recurrent neural network (RNN) designed to retain long-term dependencies in sequential data, making it effective for tasks involving time-series or sequential information.
	
	\item \textbf{Gated Recurrent Unit (GRU):} Similar to LSTM, GRU is another variant of RNN designed to address the vanishing gradient problem. It simplifies the architecture while maintaining the capability to capture dependencies in sequential data.
	
	\item \textbf{Hybrid LSTM-GRU Model:} This model combines elements of both LSTM and GRU architectures, leveraging their respective strengths to enhance overall performance.
	
	\item \textbf{Highway LSTM Model:} The Highway LSTM model incorporates highway networks with LSTM units, allowing for a more controlled flow of information through the network. This can be beneficial in learning hierarchical representations.
	
	\item \textbf{Transformer Model:} The Transformer model is a non-recurrent architecture that utilises self-attention mechanisms to capture dependencies across different positions in the input sequence simultaneously. It has demonstrated remarkable success in various natural language processing tasks.
	
\end{enumerate}

\subsection{Comparison Models}

We conduct a comprehensive comparison of our proposed models, DPE and PaDPE, against several state-of-the-art models, encompassing both classical and contemporary approaches. The aim is to evaluate the effectiveness of our models in capturing complex relationships within the data. The comparison includes the following models:

\begin{enumerate}
	\item Adaptive Boosting
	\item Gradient Boosting 
	\item K-Nearest Neighbours
	\item Combined Regression Strategy (COBRA)
\end{enumerate}

This comprehensive selection of base and comparison models provides a thorough foundation for evaluating the proposed models' capabilities and understanding their potential advantages in the context of the specific tasks under consideration.

\subsection{Hyperparameter Tuning}

  In the pursuit of a rigorous and equitable comparison among various models, hyper-parameter optimization through cross-validation is employed. The hyper-parameter search space, as delineated in Table \ref{tab:HPspace}, serves as a crucial framework for this optimization process. Certain parameters remain constant throughout the optimisation, maintaining uniformity across all pertinent models. These fixed parameters include a batch size set to $32$, a learning rate fixed at $0.001$, and a fixed number of epochs at $80$.

  For models DPE and PaDPE, specific hyper-parameters undergo meticulous tuning through this process. Notably, the parameters subject to tuning encompass the distance parameter denoted as $\epsilon$, the number of machines represented by $\alpha$, and the fraction denoted as $\frac{l}{n}$. The specific range for these hyper-parameters is elucidated in Table \ref{tab:HPspace}, guiding the optimisation process to attain optimal model performance.

\begin{table}[h]
	\caption{Hyper-parameter search space for the base and proposed models.}
	\label{tab:HPspace}
	\centering
	\begin{tabular}{|l|l|l|}
		\hline
		\textbf{Model} & \textbf{Parameter} & \textbf{Values} \\
		\hline
		LSTM & nodes & $[16, 32, 50, 64, 96, 100, 128]$ \\
		& Layers & $[0, 1, 2, 3]$ \\
		& Optimizer & Adam \\
		& Activation & [ReLU, $\tanh$] \\
		& Dropout Rate & $(0, 0.5)$ \\
		\hline
		GRU & nodes & $[16, 32, 50, 64, 96, 100, 128]$ \\
		& Layers & $[0, 1, 2, 3]$ \\
		& Optimizer & Adam \\
		& Activation & [ReLU, $\tanh$] \\
		& Dropout Rate & (0, 0.5) \\
		\hline
		Hybrid LSTM & LSTM nodes & $[16, 32, 50, 64, 96, 100, 128]$ \\
		& GRU nodes & $[16, 32, 50, 64, 96, 100, 128]$ \\
		& LSTM Layers & $[0, 1, 2, 3]$ \\
		& Optimizer & Adam \\
		& Activation & [ReLU, $\tanh$, sigmoid] \\
		& Dropout Rate & $(0, 0.5)$ \\
		\hline
		Highway LSTM & LSTM nodes & $[16, 32, 50, 64, 96, 100, 128]$ \\
		& Layers & $[1, 2, 3, 4, 5]$ \\
		& t\_bias & $(-5, 5)$\\
		& Optimizer & Adam \\
		& acti\_h & ReLU \\
		& acti\_t & sigmoid \\
		& learning rate & $(1e-6, 1e-2)$ \\
		\hline
		Transformer & nodes & $(32, 200, 2)$ \\
		& Layers & $[1, 2, 3, 4, 5]$ \\
		& Optimizer & Adam \\
		& Activation & [ReLU, $\tanh$, sigmoid] \\
		& d\_k / d\_v & $[32, 64, 96]$ \\
		& learning rate & $(1e-5, 1e-2)$ \\
		& Dropout Rate & $(0, 0.5)$ \\
		& feedforward dimension & $(32, 200, 2)$ \\
		& Number of heads & $[1,2,4,8,12]$\\
		\hline
		DPE-based models & $\epsilon$ & $(0,1)$ \\
		& $\alpha$ & [$\frac{1}{5}$, $\frac{2}{5}$, $\frac{3}{5}$, $\frac{4}{5}$, $1$] \\
		& $\frac{n_1}{n}$ & $(0,1)$\\
		\hline
	\end{tabular}
\end{table}

\begin{table}[h]
	\caption{Hyper-parameter search space for the benchmark models.}
	\label{tab:HPspace2}
	\centering
	\begin{tabular}{|l|l|l|}
		\hline
		\textbf{Model} & \textbf{Parameter} & \textbf{Values} \\
		\hline
		AdaBoost & estimators & $[10, 50, 100, 200, 300, 400, 500]$ \\
		& learning Rate & $[10^{-3}, 10^{-2}, 10^{-1}]$ \\
		\hline
		XGBoost& estimators & $[10, 50, 100, 200, 300, 400, 500]$ \\
		& depth & $[1, 2, 3, 4, 5]$ \\
		& subsample & $[0.5, 0.6, 0.7, 0.8, 0.9]$ \\
		& child weight & $[2, 4, 6, 8, 10]$ \\
		\hline
    	  K-NN & neighbours & $[1, 2, 3, 4, 5]$ \\
    		& weights & $[\text{uniform}, \text{distance}]$ \\
            & P & $[1, 2, 3, 4, 5]$ \\
            \hline
            COBRA & $\epsilon$ & $(0,1)$ \\
    		& $\alpha$ & [$\frac{1}{5}$, $\frac{2}{5}$, $\frac{3}{5}$, $\frac{4}{5}$, $1$] \\
    		& $\frac{l}{n}$ & $(0,1)$\\
            \hline
	\end{tabular}
\end{table}

\section{Empirical Results and Discussion}

\subsection{Performance evaluation over multivariate time series forecasting}

\begin{figure*}[!h]
	\centering
	\includegraphics[scale=0.3]{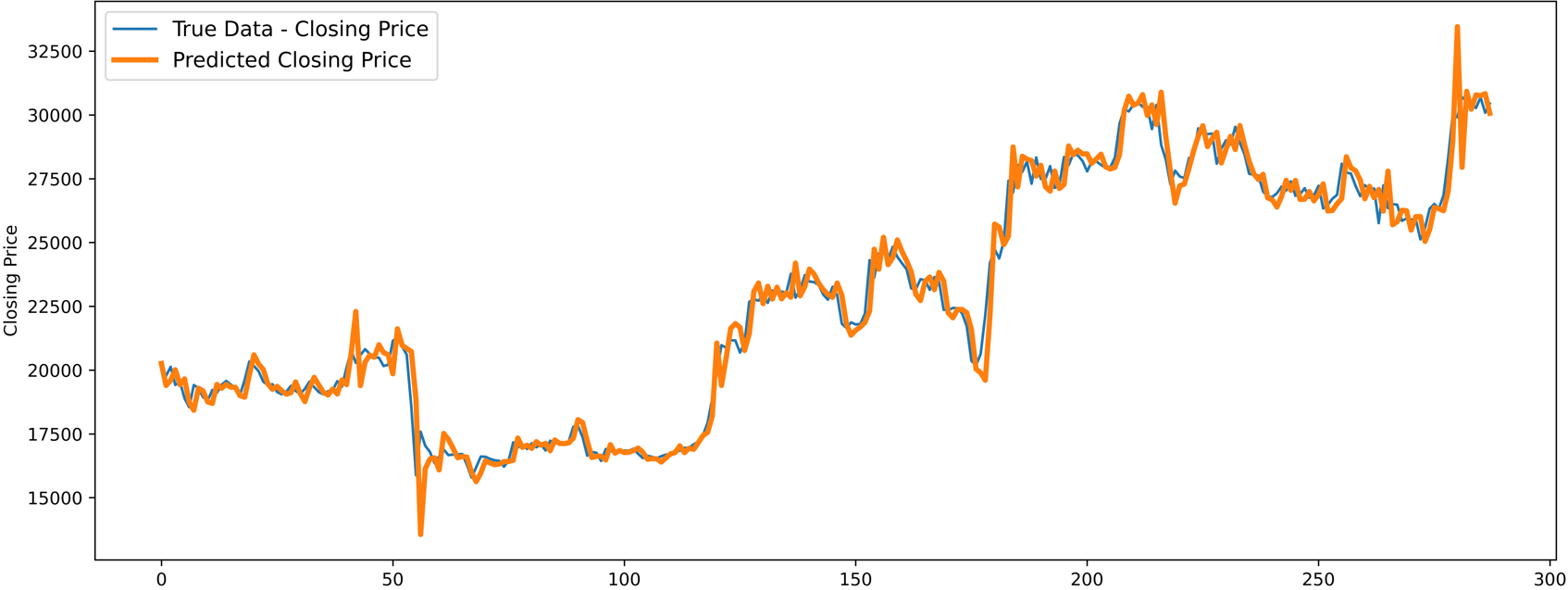}
	\caption{The visualization illustrates the Closing Price of the Bitcoin dataset employing the proposed DPE methodology.}
	\label{fig:cp}
\end{figure*}

\begin{figure*}[!h]
	\centering
	\includegraphics[scale=0.27]{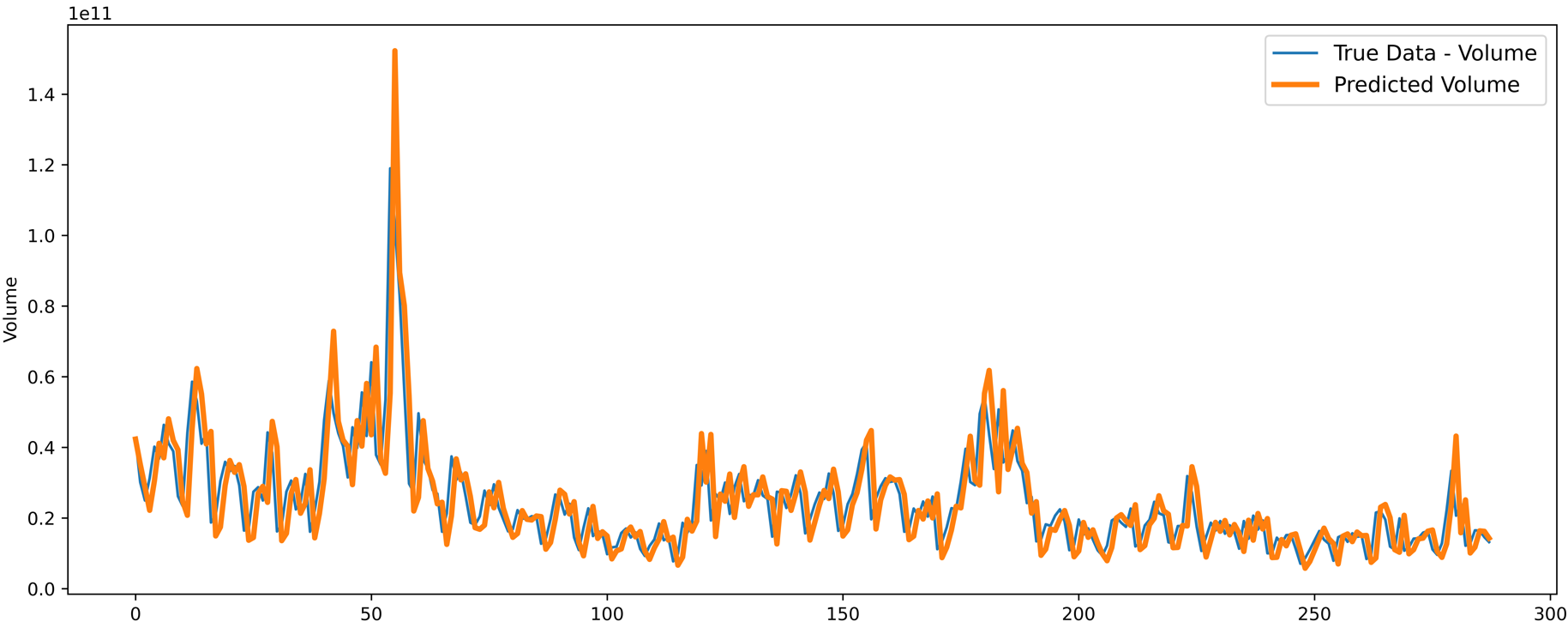}
	\caption{The visualization illustrates the Volume of the Bitcoin dataset employing the proposed DPE methodology.}
	\label{fig:volume}
\end{figure*}

  Tables \ref{table:rmse} and \ref{table:mase} provide a detailed comparison of different forecasting models applied to distinct time series, using metrics like Root Mean Square Error (RMSE) and Mean Absolute Percentage Error (MAPE). Bold highlights in the tables indicate instances where a specific model excels for a given time series. Importantly, the DPE model consistently shows exceptional accuracy across most datasets. While the proposed PaDPE model outperforms others in certain datasets, there is noticeable variability in its results. Additionally,  XGBoost ensemble models consistently demonstrate outstanding accuracy compared to their counterparts, highlighting their strong predictive capabilities.

  In this research, we apply the proposed models on a bitcoin finance dataset. A visual representations of the raw price and volume of the dataset are available in Figures \ref{fig:cp} and \ref{fig:volume} respectively.  The figures also present a side-by-side view of DPE predictions on respective test datasets, providing a detailed understanding of the model's predictive strengths. Specifically, the closing and volume prices for the Bitcoin dataset were meticulously studied, shedding light on the model's effectiveness in capturing intricate patterns within financial time series data.

\begin{sidewaystable}
\centering
\caption{RMSE results for multivariate time series forecasting}
\label{table:rmse}
    \begin{adjustbox}{width=\textwidth}
        \begin{tabular}{|c|c|c|c|c|c|c|c|c|c|c|c|c|}
            \hline
            Location & Month & LSTM & GRU & Hybrid-LSTM & Highway-LSTM & Transformer & AdaBoost & XGBoost & k-NN & COBRA &  DPE & PaDPE \\ 
            \hline
            & Bitcoin & $0.0136$ & $0.0135$ & $0.0134$ & $0.0139$ & $0.0132$ & $0.0123$ & $\textbf{0.0119}$ & $0.0132$ & $0.0179$ &  $0.0125$ & $0.0127$ \\
            Finance & DJI & $0.0089$ & $0.0087$ & $0.0093$ & $0.0094$ & $0.0092$ & $0.0086$ & $\textbf{0.0083}$ & $0.0093$ & $0.0098$ &  $0.0087$ & $0.0086$ \\
            & S\&P500 & $0.0069$ & $0.0073$ & $0.0073$ & $0.0087$ & $0.0071$ & $0.0062$ & $0.0062$ & $0.0072$ & $0.0072$ &  $\textbf{0.006}$ & $\textbf{0.006}$ \\
            \hline

            & SA & $0.0141$ & $0.0184$ & $\textbf{0.0078}$ & $0.0183$ & $0.0110$ & $0.0126$ & $0.0103$ & $0.0309$ & $0.0162$ &  $\textbf{0.0078}$ & $0.0082$ \\
            & NSW & $0.0071$ & $0.0071$ & $0.0078$ & $0.0083$ & $0.0098$ & $0.0545$ & $0.0503$ & $0.0228$ & $0.0090$ &  $0.0070$ & $\textbf{0.0069}$ \\
            Load & VIC & $0.0010$ & $0.0091$ & $0.0092$ & $\textbf{0.0089}$ & $0.0137$ & $0.0134$ & $0.0102$ & $0.0312$ & $0.0097$ &  $0.0101$ & $0.0101$ \\
            & TAS & $0.0124$ & $0.0123$ & $0.0082$ & $0.0084$ & $0.0103$ & $0.0188$ & $0.0140$ & $0.0202$ & $0.0086$ &  $\textbf{0.0076}$ & $0.0077$ \\
            & QLD & $0.0037$ & $0.0035$ & $0.0035$ & $0.0045$ & $0.0040$ & $0.0043$ & $0.0669$ & $0.0415$ & $0.00401$ &  $\textbf{0.003}$ & $\textbf{0.003}$ \\
            \hline
        \end{tabular}
    \end{adjustbox}
\end{sidewaystable}

\begin{sidewaystable}[h]
\centering
    \caption{MAPE results for multivariate time series forecasting}
    \label{table:mase}
    \begin{adjustbox}{width=\textwidth}
            \begin{tabular}{|c|c|c|c|c|c|c|c|c|c|c|c|c|}
                \hline
                Location & Month & LSTM & GRU & Hybrid-LSTM & Highway-LSTM & Transformer & AdaBoost & XGBoost & k-NN & COBRA &  DPE & PaDPE \\ 
                \hline
                & Bitcoin & $0.0077$ & $0.0078$ & $0.0078$ & $0.0075$ & $0.0075$ & $0.0067$ & $\textbf{0.0062}$ & $0.0074$ & $0.0087$ &  $0.0072$ & $0.0072$ \\
                Finance & DJI & $0.0045$ & $0.0044$ & $0.0046$ & $0.0049$ & $0.0046$ & $0.0043$ & $\textbf{0.0038}$ & $0.0045$ & $0.0065$ &  $0.0042$ & $0.0042$ \\
                & S\&P500 & $0.0036$ & $0.0037$ & $0.0038$ & $0.0041$ & $0.0039$ & $0.0031$ & $0.0031$ & $0.0037$ & $0.0039$ &  $\textbf{0.003}$ & $\textbf{0.003}$ \\
                \hline
    
                & SA & $0.2163$ & $0.1552$ & $\textbf{0.0896}$ & $0.1562$ & $0.1208$ & $0.099$ & $0.1007$ & $0.2082$ & $0.1457$ &  $0.0861$ & $0.091$ \\
                & NSW & $0.0050$ & $0.0049$ & $0.0052$ & $0.0.0055$ & $0.0073$ & $0.0506$ & $0.0344$ & $0.0241$ & $0.0068$ &  $0.0050$ & $\textbf{0.0048}$ \\
                Load & VIC & $0.1061$ & $0.1174$ & $0.1082$ & $0.1322$ & $0.1690$ & $0.1658$ & $0.1084$ & $0.3269$ & $0.0997$ &  $0.1333$ & $0.1334$ \\
                & TAS & $0.1350$ & $0.1343$ & $0.0667$ & $0.0702$ & $0.1227$ & $0.197$ & $0.1710$ & $0.2159$ & $0.0702$ &  $\textbf{0.0624}$ & $0.0635$ \\
                & QLD & $0.0019$ & $0.0019$ & $0.0017$ & $0.0022$ & $0.0020$ & $0.0044$ & $0.0020$ & $0.003$ & $0.0019$ &  $\textbf{0.0016}$ & $\textbf{0.0016}$ \\
                \hline
            \end{tabular}
    \end{adjustbox}
\end{sidewaystable}

\subsubsection{Statistical comparison using Wilcoxon's test}

  Despite these notable performances, selecting the most suitable forecasting method based only on error metrics is challenging due to the competitive outcomes among models. Further analysis is conducted using statistical tests to probe differences among all models. This indicates substantial differences in forecasting models' performances across the eight datasets. The use of statistical tests adds depth to the analysis beyond error metrics, providing insights into the comparative strengths and weaknesses of the forecasting models under consideration.

Table \ref{table:wtrmse} and \ref{table:wtmape} showcase the results of statistical comparisons involving the proposed models: DPE, PaDPE, and other models used for comparison. These tables display the average ranks, with the model achieving the lowest rank considered the most favourable in terms of performance. The proposed models DPE and PaDPE consistently attains a lower average rank across all tables, as illustrated in Tables \ref{table:wtrmse} and \ref{table:wtmape}.

\begin{table*}[!htbp]
    \centering
    \caption{Statistical comparison between proposed and other models over RMSE.}
    \label{table:wtrmse}
    \begin{tabular}{|c|c|c|}
        \hline
        {Method} & {Avg. Rank} & {p-value} \\
        \hline
        $\text{DPE}^{\dagger}$ & $\textbf{2.94}$ & \\
        $\text{PaDPE}^{\dagger}$ & $3.000$ & $2e-1$\\
        COBRA & $8.50$ & $1e-2$\\
        K-NN & $9.94$ & $7e-3$\\
        XGBoost & $6.50$ & $1e-1$\\
        AdaBoost & $7.13$ & $5e-2$\\
        Transformer & $7.31$ & $7e-3$\\
        Highway-LSTM & $8.25$ & $3e-2$\\
        Hybrid-LSTM & $5.56$ & $1e-1$\\
        GRU & $6.63$ & $7e-2$\\
        LSTM & $5.94$ & $1e-1$\\        
        \hline 
    \end{tabular}
\end{table*}

\begin{table*}[!htbp]
    \centering
    \caption{Statistical comparison between proposed and other models over MAPE.}
    \label{table:wtmape}
    \begin{tabular}{|c|c|c|}
        \hline
        {Method} & {Avg. Rank} & {p-value} \\
        \hline
        $\text{DPE}^{\dagger}$ & $\textbf{2.88}$ & \\
        $\text{PaDPE}^{\dagger}$ & $2.94$ & $5e-1$\\
        COBRA & $7.75$ & $1e-1$\\
        K-NN & $9.25$ & $7e-3$\\
        XGBoost & $5.63$ & $4e-1$\\
        AdaBoost & $7.31$ & $3e-2$\\
        Transformer & $8.38$ & $7e-3$\\
        Highway-LSTM & $8.50$ & $7e-2$\\
        Hybrid-LSTM & $5.38$ & $1e-1$\\
        LSTM & $6.50$ & $1e-1$\\
        GRU & $6.63$ & $5e-1$\\        
        \hline 
    \end{tabular}
\end{table*}

\subsection{Abliation Study}

\subsubsection{Comparative analysis of Two Tuning Methods : BOA and Grid based tuning}

  The suggested model comprises two crucial elements: BOA hyper-parameter tuning and the \textbf{proposed methodologies}, DPE and PaDPE. To delve into the significance of each component, we conduct an ablation study. This entails the formulation and assessment of five variants, each representing a distinct combination of the model's components. These variants are systematically evaluated across diverse datasets, enabling us to gauge the necessity and impact of each element. The overall performance is then averaged across all datasets, and errors are normalised to accentuate variations among them. The five variants under examination are:

\begin{itemize}
        \item GridCOBRA: The DPE utilises grid search for parameter optimisation.
	\item BOACOBRA: The combined regression strategy utilises BOA to optimise parameters in original COBRA.
	\item GridDPE: The DPE utilises grid search for parameter optimisation.
	\item BOADPE: The DPE utilises both the BOA for parameter optimisation.
	\item BOAPaDPE: The PaDPE utilises both the BOA for parameter optimisation.
	\item GridPaDPE: The PaDPE utilises gridserach for parameter optimisation.
	\end{itemize}

\begin{figure*}[!ht]
	\centering
	\includegraphics[scale=0.6]{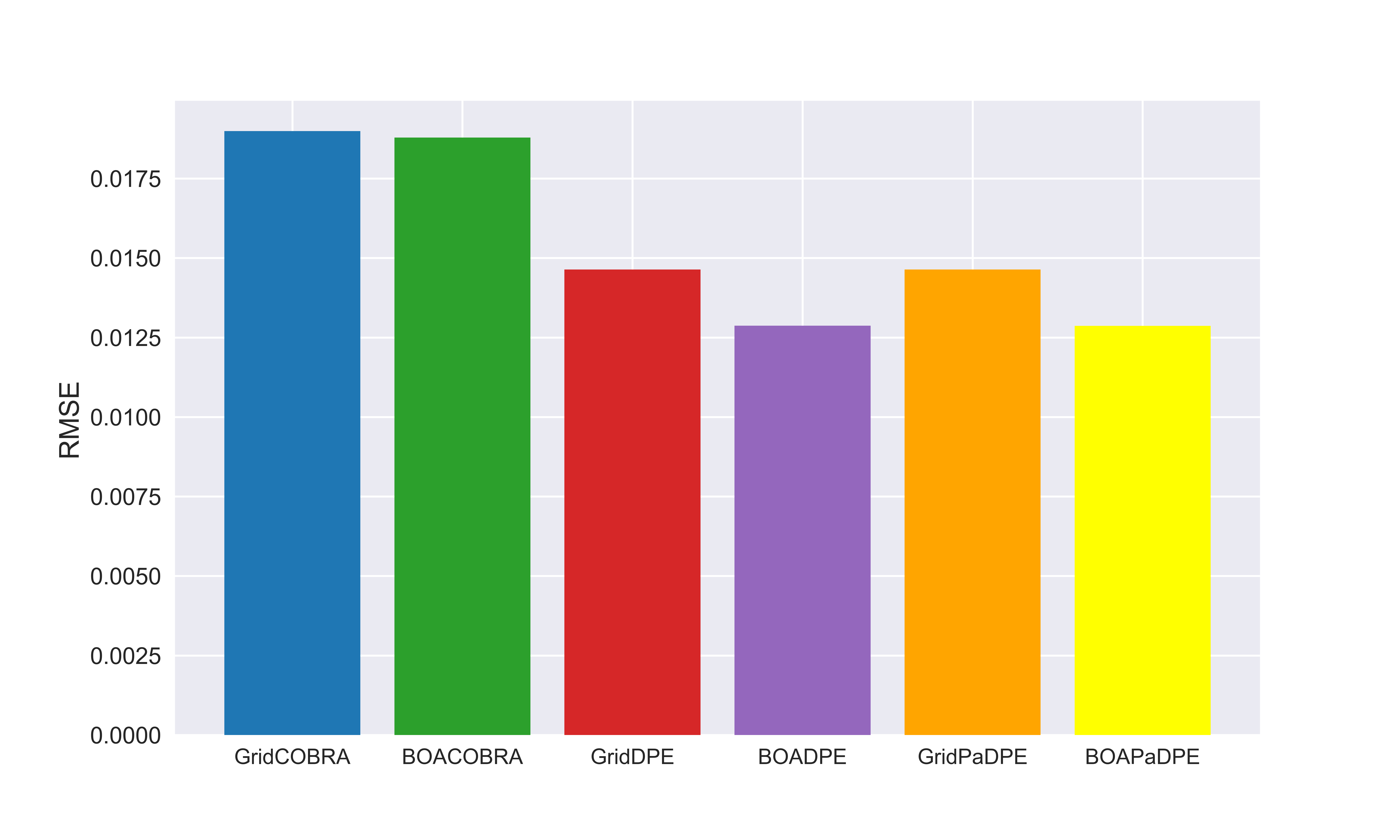}
	\caption{The bar-plot of the performance of different variations of COBRA with respect to RMSE}
	 \label{fig:rmseboa}
\end{figure*}

\begin{figure*}[!ht]
	\centering
	\includegraphics[scale=0.6]{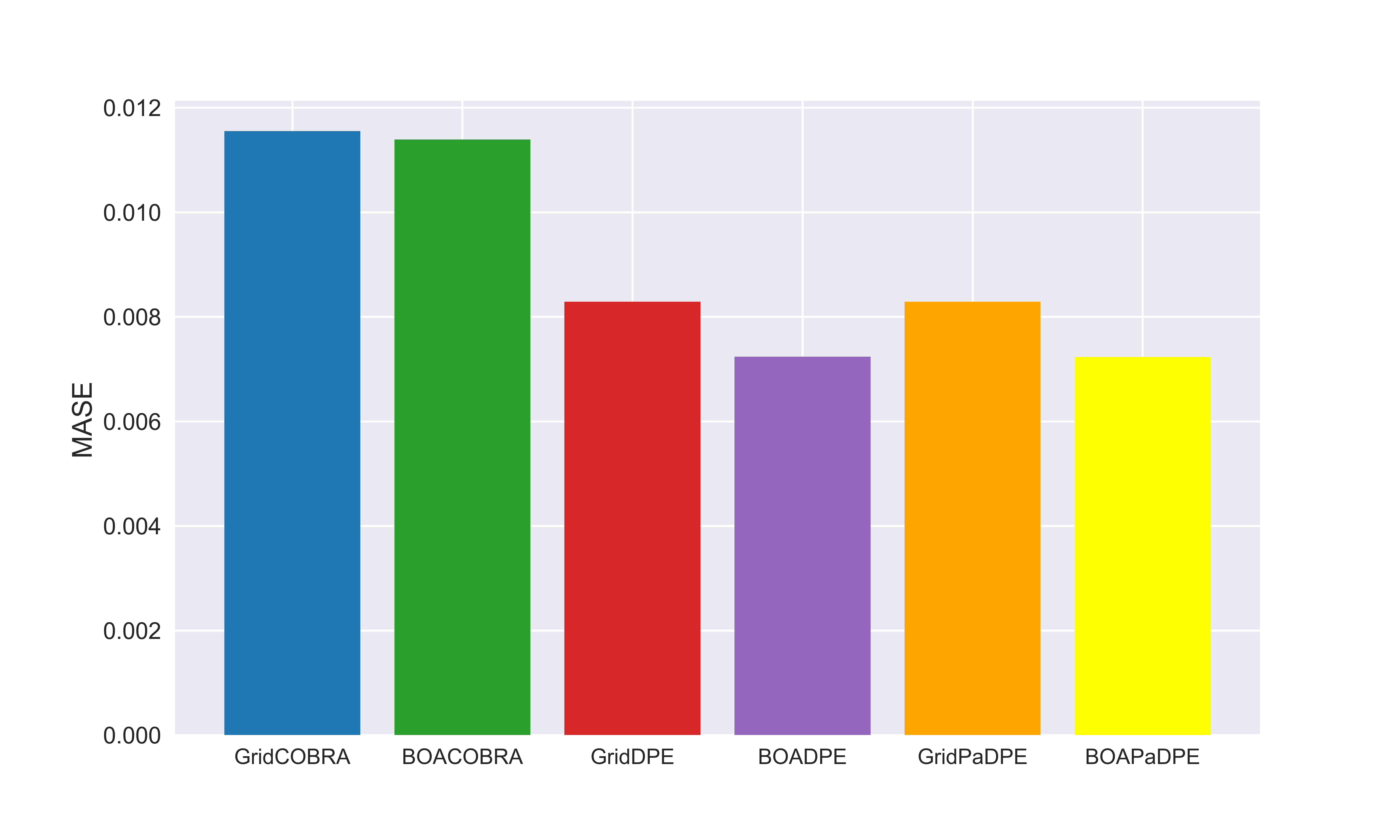}
	\caption{The bar-plot of the performance of different variations of COBRA with respect to MAPE}
	\label{fig:maseboa}
\end{figure*}

   The bar-plots shows the performance of different variations of COBRA in sequential prediction setup with respect to RMSE in Figure-\ref{fig:rmseboa} and Figure-\ref{fig:maseboa} respectively.   The above comparison between the proposed BOA methodologies and their grid-based counterparts reveals that BOA based models consistently outperform the Grid-search based models.  This underlines the robustness and effectiveness of the BOA methodologies in enhancing the study's overall outcomes.  Moreover, BOADPE outperforms all models in both the metrics considered in this paper.

\subsection{Some Analysis on RMSE and Model Parameters}

  We adapt a sensitivity analysis for both the proposed methodologies, namely DPE and PaDPE and conduct a simulation study by systematically varying the number of machines represented by the parameter $\alpha$, while maintaining a constant optimal value for $\epsilon$. Subsequently, we varied the parameter $\epsilon$ while keeping $\alpha$ fixed.

  These approaches allow us to assess the robustness and performance of both DPE and PaDPE under different conditions. By systematically altering key parameters, namely $\alpha$ and $\epsilon$, we gained insights into how these variations impact the outcomes of the simulation studies.

\subsubsection{Varying the number of machines}

\begin{figure*}[!ht]
	\centering
	\includegraphics[scale=0.7]{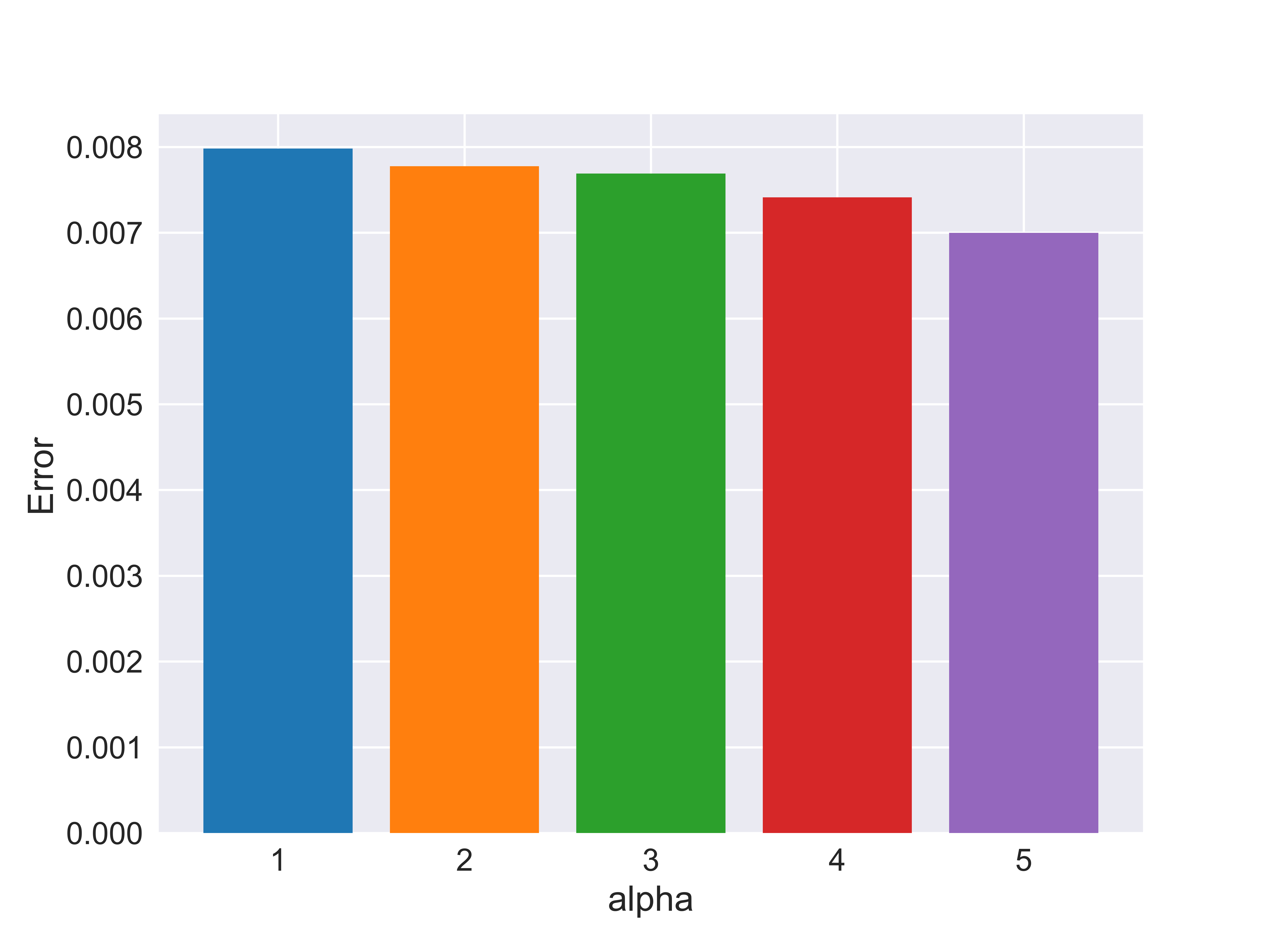}
	\caption{This barplot illustrates the relation between RMSE and the number of models}
	\label{fig:salpha}
\end{figure*}

  In Figure-\ref{fig:salpha}, we systematically explore the interplay between machine numbers, denoted as alpha, and the Mean Squared Error (MSE).  We vary alpha values within the range of $1$ to $5$ and observe a consistent reduction in the Mean Squared Error providing optimal $\alpha$ at 5 for DJI dataset.  The above study also indicates that more number of models together improve DPE accuracy, ultimately resulting in a noteworthy reduction in errors.
  
\subsubsection{Varying the distance parameter $\epsilon$}

\begin{figure*}[!ht]
	\centering
	\includegraphics[scale=0.7]{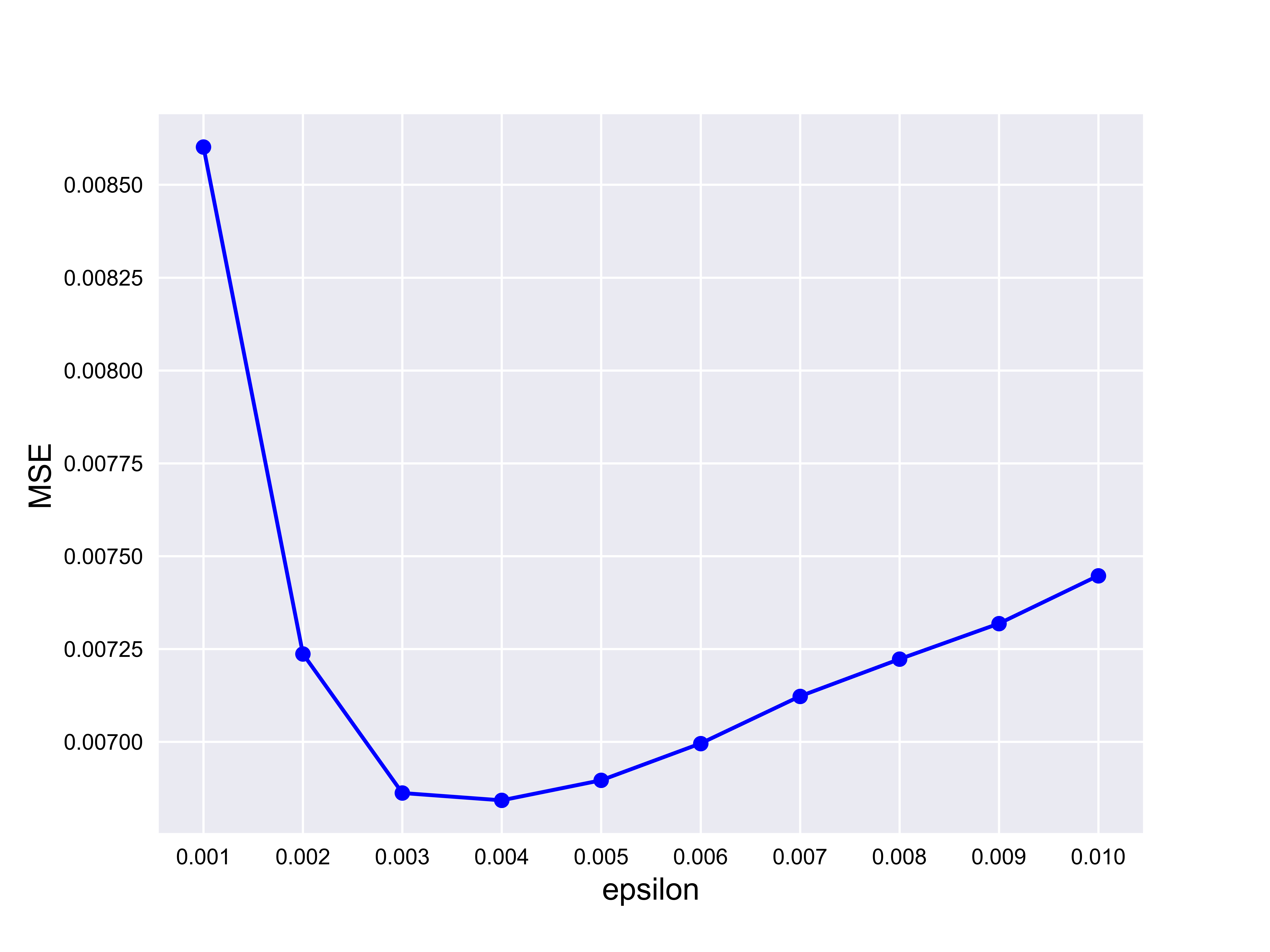}
	\caption{This relation between the MSE and the distance parameter $\epsilon$}
	\label{fig:sepsilon}
\end{figure*}

  We also explore the relation between the distance parameter epsilon and Mean Squared Error (MSE). The study involves a systematic variation of epsilon values, ranging from 0.001 to 0.01.  We keep the alpha parameter at a constant value.   Figure-\ref{fig:sepsilon} shows the curve for MSE with respect to $\epsilon$ convex in nature, finally providing the minimum value of $\epsilon$ at $0.004$.

  Our findings provide valuable information on the sensitivity of the proposed methodologies to changes in the number of machines ($\alpha$) and the threshold parameter, $\epsilon$. This comprehensive analysis enhances our understanding of the behaviour and effectiveness of DPE and PaDPE, contributing to the refinement of these methodologies in all other datasets.

\section{Dynamic Prediction}

  We assess our model's effectiveness through dynamic predictions, wherein the model operates through dynamically changing scalers.  The scaler takes step-by-step update and we fed the scaled data to the model.  After obtaining the predictions we inverse-scaled the result and concatenate the same with past data. Time-lagged entries help to preserve temporal context, and we define a new scaler for the next iteration. This iterative process ensures adaptability to evolving data patterns.

\begin{figure*}[!ht]
	\centering
	\includegraphics[scale=0.35]{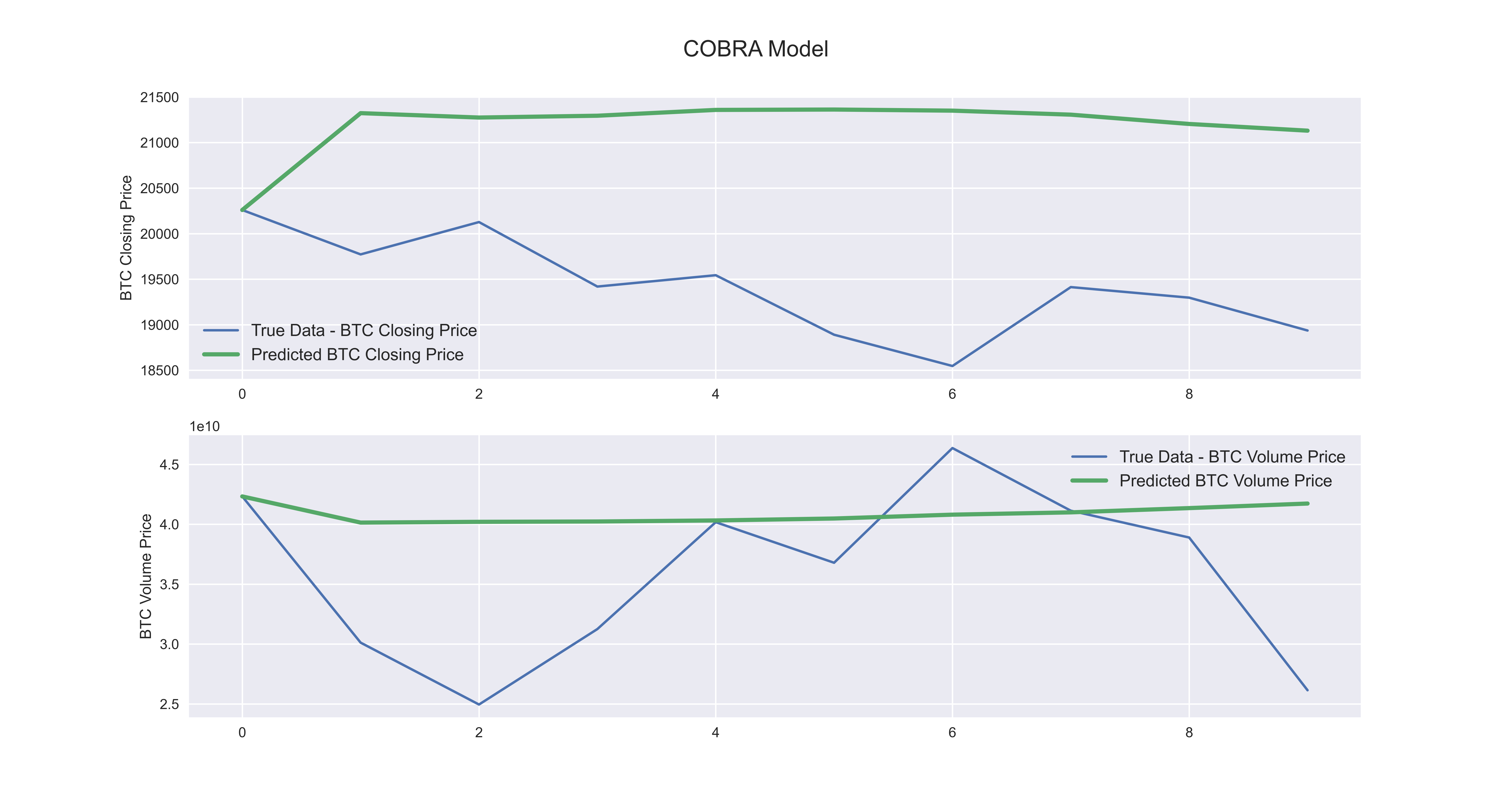}
	\caption{This figure demonstrates dynamic predictions on the Bitcoin dataset using the DPE method, with subsequent inverse scaling performed by the dynamic scaler}
	\label{fig:dynamicscaler}
\end{figure*}

  The dynamic predictions depicted in Figure-\ref{fig:dynamicscaler} showcase the future predictive capabilities of the proposed models.  These plots illustrate the predictions conducted on the dynamic scaler, with subsequent inverse scaling concerning dynamic scaler. The visualisations does not strongly emphasise the models' proficiency in dynamic predictions, since the picture is zoomed closely in ten successive prediction.  However, It captures better signal capture for volume than prices.  It depends on the practitioner how much accuracy required for certain purpose.  We need more improvement and better design in this respect.


\section{Conclusion}

  Dynamic-Based Proximity Ensemble Method and Partitioned Dynamic-Based Proximity Ensemble Method are two proposed variations which exhibit superior performance over a state-of-the-art comparative models. The efficacy of our proposed methodology is further validated through its ability to capture complex temporal patterns across diverse datasets, including cryptocurrency, stock index, and short-term load forecasting. This research contributes to the advancement of forecasting techniques in an arbitrary multidimensional signals. More improvement of the work is necessary specifically in capturing accurate evolving future data patterns. Additionally, exploring adaptive combinations for the final prediction could enhance the flexibility and robustness of the ensemble method.  An interval estimation in this context could be a challenging problem. The work is in progress.

\section*{Declarations}

\begin{itemize}
	\item Funding: The work is not related to any funding.
	
	\item Conflict of interest/Competing interests: This work has no conflict of interest or competing interests.
	
	\item Ethics approval: The paper does not have any ethics-related issues.
	
	\item Consent to participate: All co-authors provide consent to participate in this research article. 
	
	\item Consent for publication: All co-authors have consented to keep their names in related publications.
	
	\item Availability of data and materials: All datasets are publicly available and downloaded directly from Yahoo Finance and AEMO.
	
	\item Code availability: The codes are available to each author. We can make it accessible through publicly available through Github link on request.
	
	\item Authors' contributions: Aryan Bhambu is the first author of the paper.  He was associated with all computation, empirical study, proposed model creation, and manuscript completion. Arabin Kumar Dey is the corresponding author and mentor for the project.
\end{itemize}

\bibliographystyle{elsarticle-harv} 
\bibliography{example}

\begin{thebibliography}{23}
\expandafter\ifx\csname natexlab\endcsname\relax\def\natexlab#1{#1}\fi
\providecommand{\url}[1]{\texttt{#1}}
\providecommand{\href}[2]{#2}
\providecommand{\path}[1]{#1}
\providecommand{\DOIprefix}{doi:}
\providecommand{\ArXivprefix}{arXiv:}
\providecommand{\URLprefix}{URL: }
\providecommand{\Pubmedprefix}{pmid:}
\providecommand{\doi}[1]{\href{http://dx.doi.org/#1}{\path{#1}}}
\providecommand{\Pubmed}[1]{\href{pmid:#1}{\path{#1}}}
\providecommand{\bibinfo}[2]{#2}
\ifx\xfnm\relax \def\xfnm[#1]{\unskip,\space#1}\fi
\bibitem[{Abbasimehr and Paki(2021)}]{abbasimehr2021prediction}
\bibinfo{author}{Abbasimehr, H.}, \bibinfo{author}{Paki, R.},
  \bibinfo{year}{2021}.
\newblock \bibinfo{title}{Prediction of covid-19 confirmed cases combining deep
  learning methods and bayesian optimization}.
\newblock \bibinfo{journal}{Chaos, Solitons \& Fractals} \bibinfo{volume}{142},
  \bibinfo{pages}{110511}.
\bibitem[{Alizadeh et~al.(2021)Alizadeh, Bafti, Kamangir, Zhang, Wright and
  Franz}]{alizadeh2021novel}
\bibinfo{author}{Alizadeh, B.}, \bibinfo{author}{Bafti, A.G.},
  \bibinfo{author}{Kamangir, H.}, \bibinfo{author}{Zhang, Y.},
  \bibinfo{author}{Wright, D.B.}, \bibinfo{author}{Franz, K.J.},
  \bibinfo{year}{2021}.
\newblock \bibinfo{title}{A novel attention-based lstm cell post-processor
  coupled with bayesian optimization for streamflow prediction}.
\newblock \bibinfo{journal}{Journal of Hydrology} \bibinfo{volume}{601},
  \bibinfo{pages}{126526}.
\bibitem[{Biau et~al.(2016)Biau, Fischer, Guedj and Malley}]{biau2016cobra}
\bibinfo{author}{Biau, G.}, \bibinfo{author}{Fischer, A.},
  \bibinfo{author}{Guedj, B.}, \bibinfo{author}{Malley, J.D.},
  \bibinfo{year}{2016}.
\newblock \bibinfo{title}{Cobra: A combined regression strategy}.
\newblock \bibinfo{journal}{Journal of Multivariate Analysis}
  \bibinfo{volume}{146}, \bibinfo{pages}{18--28}.
\bibitem[{Chung et~al.(2014)Chung, Gulcehre, Cho and
  Bengio}]{chung2014empirical}
\bibinfo{author}{Chung, J.}, \bibinfo{author}{Gulcehre, C.},
  \bibinfo{author}{Cho, K.}, \bibinfo{author}{Bengio, Y.},
  \bibinfo{year}{2014}.
\newblock \bibinfo{title}{Empirical evaluation of gated recurrent neural
  networks on sequence modeling}.
\newblock \bibinfo{journal}{arXiv preprint arXiv:1412.3555} .
\bibitem[{De~Livera et~al.(2011)De~Livera, Hyndman and
  Snyder}]{de2011forecasting}
\bibinfo{author}{De~Livera, A.M.}, \bibinfo{author}{Hyndman, R.J.},
  \bibinfo{author}{Snyder, R.D.}, \bibinfo{year}{2011}.
\newblock \bibinfo{title}{Forecasting time series with complex seasonal
  patterns using exponential smoothing}.
\newblock \bibinfo{journal}{Journal of the American statistical association}
  \bibinfo{volume}{106}, \bibinfo{pages}{1513--1527}.
\bibitem[{Fan et~al.(2018)Fan, Wang, Wu, Zhou, Zhang, Yu, Lu and
  Xiang}]{fan2018comparison}
\bibinfo{author}{Fan, J.}, \bibinfo{author}{Wang, X.}, \bibinfo{author}{Wu,
  L.}, \bibinfo{author}{Zhou, H.}, \bibinfo{author}{Zhang, F.},
  \bibinfo{author}{Yu, X.}, \bibinfo{author}{Lu, X.}, \bibinfo{author}{Xiang,
  Y.}, \bibinfo{year}{2018}.
\newblock \bibinfo{title}{Comparison of support vector machine and extreme
  gradient boosting for predicting daily global solar radiation using
  temperature and precipitation in humid subtropical climates: A case study in
  china}.
\newblock \bibinfo{journal}{Energy conversion and management}
  \bibinfo{volume}{164}, \bibinfo{pages}{102--111}.
\bibitem[{Fathi(2019)}]{fathi2019time}
\bibinfo{author}{Fathi, O.}, \bibinfo{year}{2019}.
\newblock \bibinfo{title}{Time series forecasting using a hybrid arima and lstm
  model}.
\newblock \bibinfo{journal}{Velvet Consulting} , \bibinfo{pages}{1--7}.
\bibitem[{Fu et~al.(2016)Fu, Zhang and Li}]{fu2016using}
\bibinfo{author}{Fu, R.}, \bibinfo{author}{Zhang, Z.}, \bibinfo{author}{Li,
  L.}, \bibinfo{year}{2016}.
\newblock \bibinfo{title}{Using lstm and gru neural network methods for traffic
  flow prediction}, in: \bibinfo{booktitle}{2016 31st Youth academic annual
  conference of Chinese association of automation (YAC)},
  \bibinfo{organization}{IEEE}. pp. \bibinfo{pages}{324--328}.
\bibitem[{Hochreiter and Schmidhuber(1997)}]{hochreiter1997long}
\bibinfo{author}{Hochreiter, S.}, \bibinfo{author}{Schmidhuber, J.},
  \bibinfo{year}{1997}.
\newblock \bibinfo{title}{Long short-term memory}.
\newblock \bibinfo{journal}{Neural computation} \bibinfo{volume}{9},
  \bibinfo{pages}{1735--1780}.
\bibitem[{Islam and Hossain(2021)}]{islam2021foreign}
\bibinfo{author}{Islam, M.S.}, \bibinfo{author}{Hossain, E.},
  \bibinfo{year}{2021}.
\newblock \bibinfo{title}{Foreign exchange currency rate prediction using a
  gru-lstm hybrid network}.
\newblock \bibinfo{journal}{Soft Computing Letters} \bibinfo{volume}{3},
  \bibinfo{pages}{100009}.
\bibitem[{Juditsky and Nemirovski(2000)}]{juditsky2000functional}
\bibinfo{author}{Juditsky, A.}, \bibinfo{author}{Nemirovski, A.},
  \bibinfo{year}{2000}.
\newblock \bibinfo{title}{Functional aggregation for nonparametric regression}.
\newblock \bibinfo{journal}{The Annals of Statistics} \bibinfo{volume}{28},
  \bibinfo{pages}{681--712}.
\bibitem[{Ma et~al.(2020)Ma, Ding, Cheng, Jiang, Gan and Xu}]{ma2020lag}
\bibinfo{author}{Ma, J.}, \bibinfo{author}{Ding, Y.}, \bibinfo{author}{Cheng,
  J.C.}, \bibinfo{author}{Jiang, F.}, \bibinfo{author}{Gan, V.J.},
  \bibinfo{author}{Xu, Z.}, \bibinfo{year}{2020}.
\newblock \bibinfo{title}{A lag-flstm deep learning network based on bayesian
  optimization for multi-sequential-variant pm2. 5 prediction}.
\newblock \bibinfo{journal}{Sustainable Cities and Society}
  \bibinfo{volume}{60}, \bibinfo{pages}{102237}.
\bibitem[{Mart{\'\i}nez et~al.(2019)Mart{\'\i}nez, Fr{\'\i}as, P{\'e}rez and
  Rivera}]{martinez2019methodology}
\bibinfo{author}{Mart{\'\i}nez, F.}, \bibinfo{author}{Fr{\'\i}as, M.P.},
  \bibinfo{author}{P{\'e}rez, M.D.}, \bibinfo{author}{Rivera, A.J.},
  \bibinfo{year}{2019}.
\newblock \bibinfo{title}{A methodology for applying k-nearest neighbor to time
  series forecasting}.
\newblock \bibinfo{journal}{Artificial Intelligence Review}
  \bibinfo{volume}{52}, \bibinfo{pages}{2019--2037}.
\bibitem[{Mojirsheibani(1999)}]{mojirsheibani1999combining}
\bibinfo{author}{Mojirsheibani, M.}, \bibinfo{year}{1999}.
\newblock \bibinfo{title}{Combining classifiers via discretization}.
\newblock \bibinfo{journal}{Journal of the American Statistical Association} ,
  \bibinfo{pages}{600--609}.
\bibitem[{Snoek et~al.(2012)Snoek, Larochelle and Adams}]{snoek2012practical}
\bibinfo{author}{Snoek, J.}, \bibinfo{author}{Larochelle, H.},
  \bibinfo{author}{Adams, R.P.}, \bibinfo{year}{2012}.
\newblock \bibinfo{title}{Practical bayesian optimization of machine learning
  algorithms}.
\newblock \bibinfo{journal}{Advances in neural information processing systems}
  \bibinfo{volume}{25}.
\bibitem[{Tay and Cao(2001)}]{tay2001application}
\bibinfo{author}{Tay, F.E.}, \bibinfo{author}{Cao, L.}, \bibinfo{year}{2001}.
\newblock \bibinfo{title}{Application of support vector machines in financial
  time series forecasting}.
\newblock \bibinfo{journal}{omega} \bibinfo{volume}{29},
  \bibinfo{pages}{309--317}.
\bibitem[{Vuong et~al.(2022)Vuong, Dat, Mai, Uyen et~al.}]{vuong2022stock}
\bibinfo{author}{Vuong, P.H.}, \bibinfo{author}{Dat, T.T.},
  \bibinfo{author}{Mai, T.K.}, \bibinfo{author}{Uyen, P.H.}, et~al.,
  \bibinfo{year}{2022}.
\newblock \bibinfo{title}{Stock-price forecasting based on xgboost and lstm.}
\newblock \bibinfo{journal}{Computer Systems Science \& Engineering}
  \bibinfo{volume}{40}.
\bibitem[{Wang and Chen(2021)}]{wang2021adaboost}
\bibinfo{author}{Wang, J.}, \bibinfo{author}{Chen, Y.}, \bibinfo{year}{2021}.
\newblock \bibinfo{title}{Adaboost-based integration framework coupled
  two-stage feature extraction with deep learning for multivariate exchange
  rate prediction}.
\newblock \bibinfo{journal}{Neural Processing Letters} \bibinfo{volume}{53},
  \bibinfo{pages}{4613--4637}.
\bibitem[{Yamak et~al.(2019)Yamak, Yujian and Gadosey}]{yamak2019comparison}
\bibinfo{author}{Yamak, P.T.}, \bibinfo{author}{Yujian, L.},
  \bibinfo{author}{Gadosey, P.K.}, \bibinfo{year}{2019}.
\newblock \bibinfo{title}{A comparison between arima, lstm, and gru for time
  series forecasting}, in: \bibinfo{booktitle}{Proceedings of the 2019 2nd
  international conference on algorithms, computing and artificial
  intelligence}, pp. \bibinfo{pages}{49--55}.
\bibitem[{Yang(2000)}]{yang2000combining}
\bibinfo{author}{Yang, Y.}, \bibinfo{year}{2000}.
\newblock \bibinfo{title}{Combining different procedures for adaptive
  regression}.
\newblock \bibinfo{journal}{Journal of multivariate analysis}
  \bibinfo{volume}{74}, \bibinfo{pages}{135--161}.
\bibitem[{Yang(2004)}]{yang2004aggregating}
\bibinfo{author}{Yang, Y.}, \bibinfo{year}{2004}.
\newblock \bibinfo{title}{Aggregating regression procedures to improve
  performance}.
\newblock \bibinfo{journal}{Bernoulli} \bibinfo{volume}{10},
  \bibinfo{pages}{25--47}.
\bibitem[{Zhou et~al.(2021)Zhou, Zhang, Peng, Zhang, Li, Xiong and
  Zhang}]{zhou2021informer}
\bibinfo{author}{Zhou, H.}, \bibinfo{author}{Zhang, S.}, \bibinfo{author}{Peng,
  J.}, \bibinfo{author}{Zhang, S.}, \bibinfo{author}{Li, J.},
  \bibinfo{author}{Xiong, H.}, \bibinfo{author}{Zhang, W.},
  \bibinfo{year}{2021}.
\newblock \bibinfo{title}{Informer: Beyond efficient transformer for long
  sequence time-series forecasting}, in: \bibinfo{booktitle}{Proceedings of the
  AAAI conference on artificial intelligence}, pp.
  \bibinfo{pages}{11106--11115}.
\bibitem[{Zilly et~al.(2017)Zilly, Srivastava, Koutn{\i}k and
  Schmidhuber}]{zilly2017recurrent}
\bibinfo{author}{Zilly, J.G.}, \bibinfo{author}{Srivastava, R.K.},
  \bibinfo{author}{Koutn{\i}k, J.}, \bibinfo{author}{Schmidhuber, J.},
  \bibinfo{year}{2017}.
\newblock \bibinfo{title}{Recurrent highway networks}, in:
  \bibinfo{booktitle}{International conference on machine learning},
  \bibinfo{organization}{PMLR}. pp. \bibinfo{pages}{4189--4198}.

\end{thebibliography}

\end{document}